\documentclass[10pt,twocolumn,letterpaper]{article}
\pdfoutput=1 

\usepackage{cvpr}
\usepackage{times}
\usepackage{epsfig}
\usepackage{graphicx}
\usepackage{amsmath}
\usepackage{amssymb}

\usepackage{multirow}

\usepackage{algorithm,algpseudocode}
\newcommand{\rev}[1]{#1}
\newcommand{\revb}[1]{{#1}}

\usepackage[pagebackref=true,breaklinks=true,letterpaper=true,colorlinks,bookmarks=false]{hyperref}

\cvprfinalcopy 

\newcommand{\h}{\mathbf{h}}
\newcommand{\dd}{\mathbf{d}}

\newcommand{\y}{\mathbf{y}}

\newcommand{\nj}{J}

\newcommand{\ep}{\hat{\mathbf{p}}}
\newcommand{\ex}{\hat{\mathbf{x}}}
\newcommand{\ec}{\hat{\mathbf{c}}}
\newcommand{\eza}{\hat{\mathit{z}}_{a}}

\newcommand{\eui}{\hat{u}_i}
\newcommand{\evi}{\hat{v}_i}
\newcommand{\edi}{\hat{d}_i}
\newcommand{\exi}{\hat{x}_i}
\newcommand{\eyi}{\hat{y}_i}
\newcommand{\ezi}{\hat{z}_i}

\newcommand{\fx}{f_x}
\newcommand{\fpx}{f^\prime_x}
\newcommand{\fy}{f_y}

\newcommand{\cx}{C_x}
\newcommand{\cy}{C_y}

\newcommand{\excfrom}{\ex{}^x_{c2\rightarrow{c1}}}

\DeclareMathOperator*{\argmin}{arg\,min}
\DeclareMathOperator*{\Tr}{Tr}

\begin{document}

\title{Consensus-based Optimization for 3D Human Pose Estimation in Camera Coordinates}

\author{Diogo C. Luvizon$^{1,2}$ \hspace{1cm}  Hedi Tabia$^{1,3}$  \hspace{1cm} David Picard$^{1,4}$
\vspace{0.254cm}\\
$^1$ETIS UMR 8051, Paris Seine University, ENSEA, CNRS, F-95000, Cergy, France\\
$^2$Advanced Technologies, Samsung Research Institute, Campinas, Brazil\\
$^3$IBISC, Univ. d\'{}Evry Val d\'{}Essonne, Universit\'{e} Paris Saclay\\
$^4$LIGM, UMR 8049, \'{E}cole des Ponts, UPE, Champs-sur-Marne, France\\
{\tt\small diogo.luvizon@ensea.fr}
}

\maketitle

\begin{abstract}
  3D human pose estimation is frequently seen as the task of estimating
  3D poses relative to the root body joint.
  Alternatively, we propose a 3D human pose estimation 
  method in camera coordinates, which allows effective combination of 2D
  annotated data and 3D poses \revb{and} a straightforward multi-view 
  generalization.
  To that end, we cast the problem \revb{as a view frustum space pose estimation,
  where absolute depth prediction and joint relative depth estimations are disentangled.
  Final 3D predictions are obtained in camera coordinates by the inverse camera projection.}
  Based on this, \revb{we also present a consensus-based optimization algorithm for multi-view predictions} from uncalibrated images, which requires a single monocular training procedure.
  \revb{Although our method is indirectly tied to the training camera intrinsics, it still converges for cameras with different intrinsic parameters, resulting in coherent estimations up to a scale factor.}
  Our method improves the state of the art on well known 3D human pose
  datasets, reducing the prediction error by 32\% in the most common benchmark.
  We also reported our results in absolute pose position error,
  achieving 80~mm for monocular estimations and 51~mm for multi-view, on average.
  Source code is available at \url{https://github.com/dluvizon/3d-pose-consensus}.
\end{abstract}




\section{Introduction}

\begin{figure}[!h]
  \centering
  \includegraphics[width=0.15\textwidth]{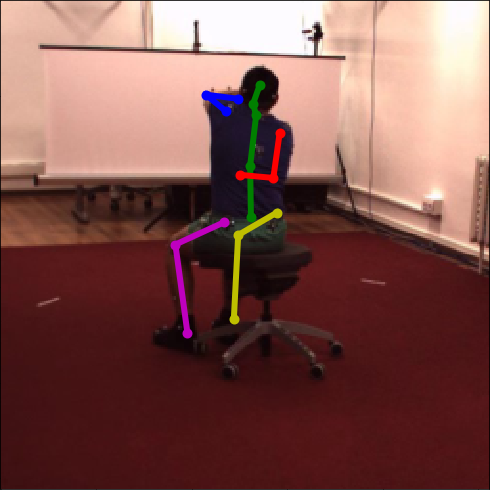}
  \includegraphics[width=0.16\textwidth]{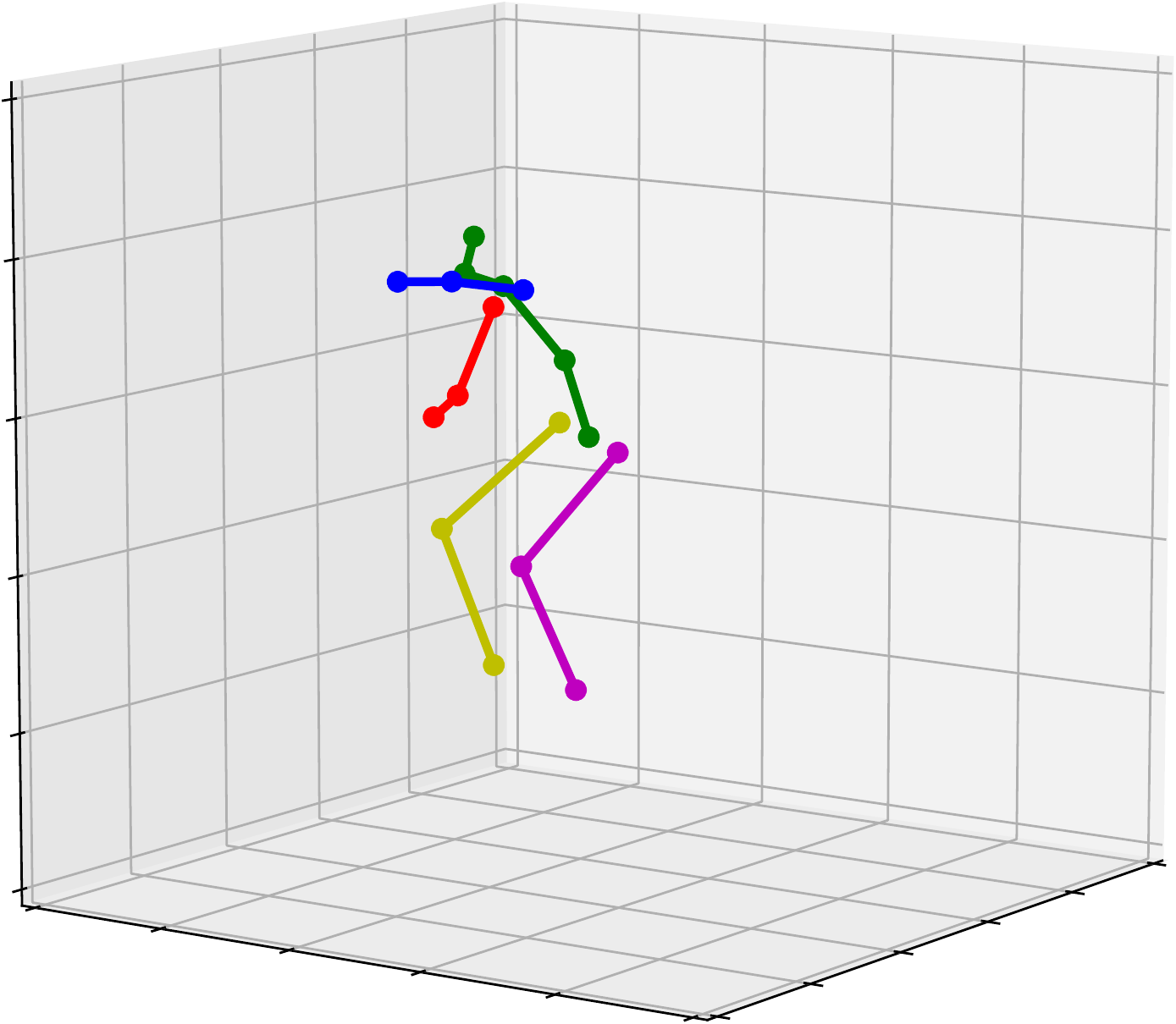}
  \includegraphics[width=0.15\textwidth]{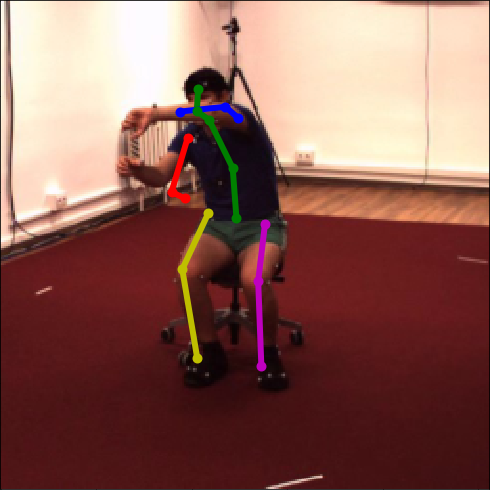}%
  \\
  \includegraphics[width=0.15\textwidth]{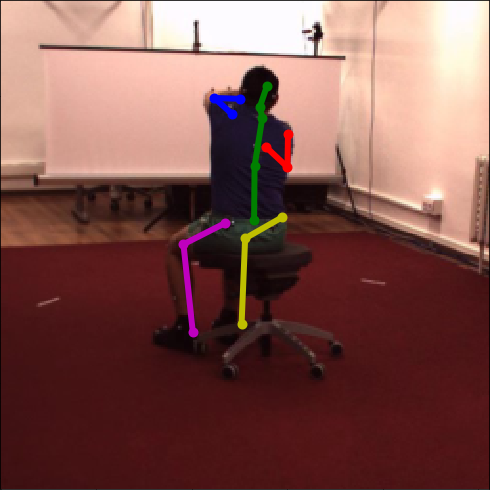}
  \includegraphics[width=0.16\textwidth]{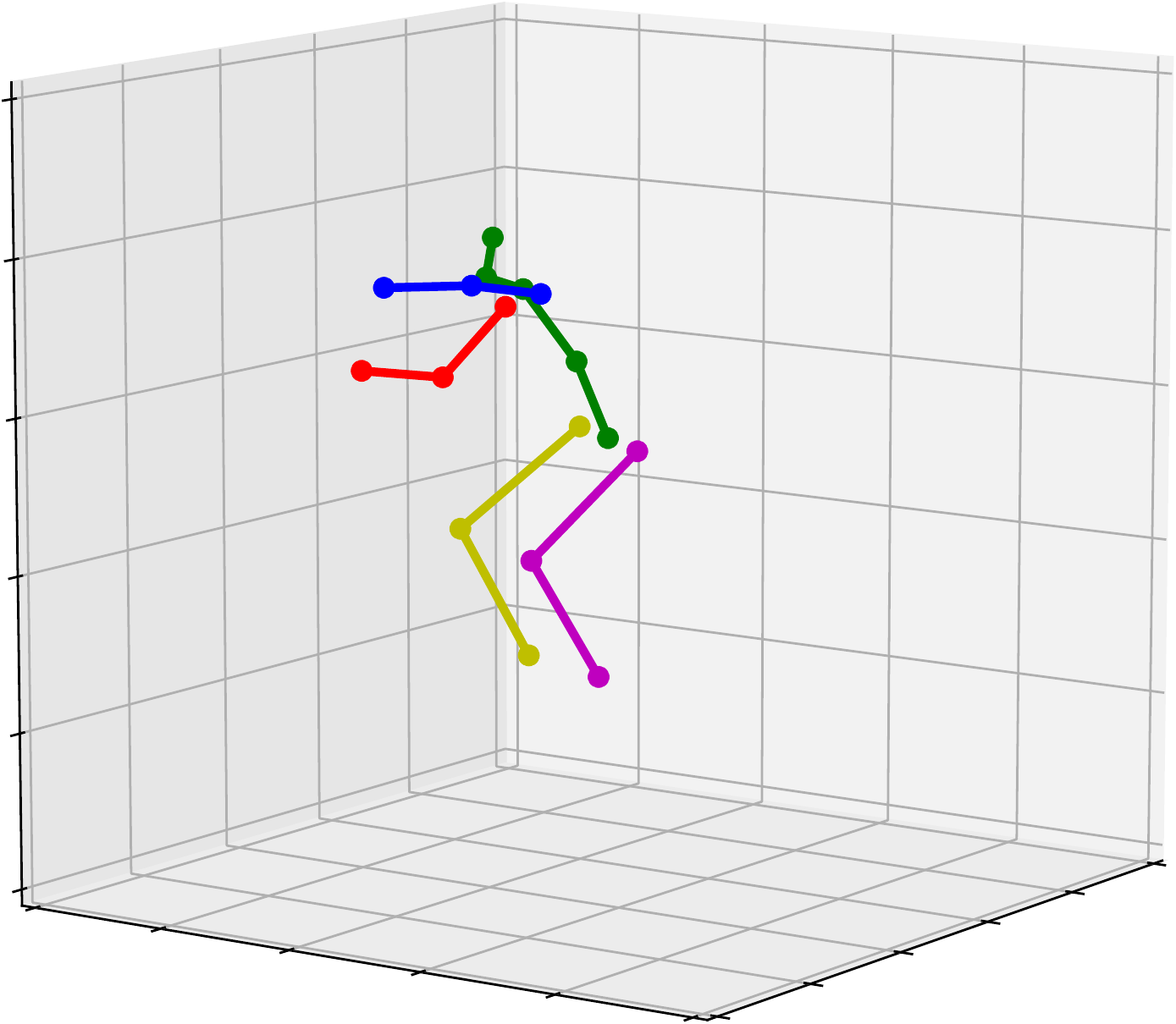}
  \includegraphics[width=0.15\textwidth]{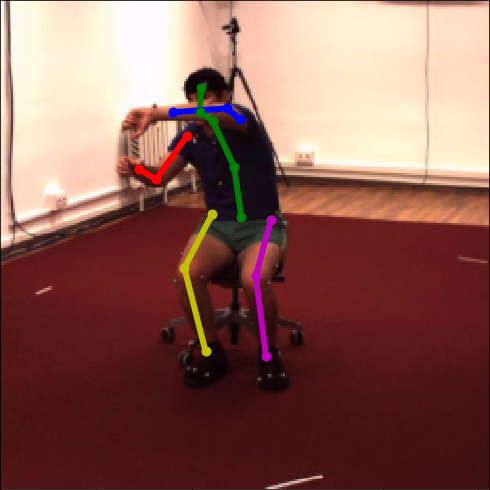}%
  \caption{Absolute 3D human pose estimated from a single image (top-left)
  with occlusion and projected into a different view (top-right). Our multi-view
  consensus-based approach (bottom) results in a more precise absolute pose
  estimation and effectively handles cases of occlusion.}
  \label{fig:intro}
\end{figure}

3D human pose estimation is a very active research topic, mainly due to the
several applications that benefit from precise human poses, such as
sports performance analysis, 3D model fitting, human behavior
understanding, among many others.
Despite the recent works on 3D human pose estimation, most of the
methods in the literature are limited to the problem of relative pose
prediction~\cite{Chen_2017_CVPR, Tome_2017_CVPR, Zhou_2016_CVPR, belagiannis20143d, amin2013multi}, where the root body joint is centered at the origin and the
remaining joints are estimated relative to the center.
This limitation hinders the generalization for multi-view scenarios
since predictions are not in the camera coordinates.
Contrarily, when estimations are relative to \revb{a static referential, predictions}
can be easily projected from one view to another,
as illustrated in Fig.~\ref{fig:intro}.

The methods in the state of the art frequently handle 3D human pose estimation
as a regression task, directly converting the input images to predicted poses in
millimeters~\cite{Sun_2018_ECCV, Li_2015_ICCV}.  However, this is a depth
learning problem, because identical distances in pixels can result in different
distances in millimeters. For example, a person close to the camera with the
hand next to the head has a distance (head to hand in mm) much shorter than a
person far from the camera with her arm extended, although both result in the
same distance in pixels.  Consequently, those methods have to learn the
intrinsic parameters indirectly.
Moreover, by predicting 3D poses directly in millimeters, the abundant
images with annotated 2D poses in pixels cannot be easily exploited, since
this data has no associated 3D information, and relative poses
predicted from one camera cannot be easily projected into a different view,
making it more difficult to handle occlusion cases in multi-view scenarios.

In our method, we tackle these
limitations by casting the problem of 3D human pose estimation into a different
perspective: instead of directly predicting pose in millimeters relative to the root joint, we predict
\revb{3D poses in the view frustum space, i.e., we predict}
$(u,v)$ coordinates in the image plane, in pixels, and the absolute depth in
millimeters. \revb{We further split depth estimation as a global absolute depth and joint relative depth estimations}.
Both 2D human pose and absolute depth estimation are well known problems
in the literature~\cite{Andriluka_2014_CVPR, NIPS2014_5539, NIPS2016_6489, laina2016deeper},
including absolute depth estimation benchmarks~\cite{Silberman:ECCV12, diode_dataset},
but are usually not correlated.
\revb{In our method, we train a feed-forward neural network by effectively merging in-the-wild 2D data and precise 3D poses, making the best use of each. Even though our network is trained only with monocular images, the predictions from individual views can be merged by the proposed consensus-based optimization in order to produce multi-view estimations, resulting in an effective way to handle the challenging cases of occlusions, as demonstrated by a significant improvement in accuracy in our experiments. Although our training scheme is indirectly tied to the camera intrinsics, our method has demonstrated a generalization capability to predict 3D poses up to a scale from a completely different camera setup, including different intrinsic parameters. This was evidenced by qualitative and quantitative evaluations.}

Considering the exposed limitations of relative 3D human pose estimation,
we aim to fill the gap of current methods by addressing the more complex
problem of absolute 3D human pose estimation, where predictions are performed
with respect to a static referential \ie, the camera position, and not to the
person's root joint.
In that direction, we present our contributions:
\textit{First}, we propose an absolute 3D human pose estimation method from
monocular cameras \revb{which achieves results in the state of the art when considering similar camera intrinsics at training and inference time}.
\textit{Second}, we propose a consensus-based optimization for multi-view
absolute 3D human pose estimation from uncalibrated images, which requires a single monocular training procedure. \revb{The multi-view estimation approach is capable of generalizing for different camera setups, resulting in coherent 3D absolute predictions up to a scale factor}.
Our method sets the new state-of-the-art results on
the challenging test set from Human3.6M, \revb{improving previous results by 10\%
with monocular predictions and by 32\% considering multiple views.}

The remaining of this paper is divided as follows. In
section~\ref{sec:relatedwork} we present the related work. Our method for 3D
human pose estimation is explained in section~\ref{sec:method-3dpose} and
our algorithm for consensus-based optimization is detailed in
section~\ref{sec:consensus-optimization}.  The experiments are presented in
section~\ref{sec:experiments} and in section~\ref{sec:conclusions} we conclude
this paper.

\section{Related work}
\label{sec:relatedwork}

In this section, we review the methods most related to our work, giving special
attention to monocular \rev{(relative and absolute)} and multi-view 3D human
pose estimation.  We recommend the survey in~\cite{SARAFIANOS20161} for readers
seeking for a more detailed review.

\subsection{Monocular relative 3D human pose estimation}

In the last decade, monocular 3D human pose estimation has been a very active
research topic in the community~\cite{Agarwal, zhou2016deep, TekinKSLF16, Li_2015_ICCV, Ionesc_ICCV_2011}.
Many recent works have proposed to directly predict relative 3D poses from
images~\cite{Sun_2018_ECCV, Sun_2017_ICCV, Pavlakos_2017_CVPR},
which requires the model to learn a complex
projection from 2D pixels to millimeters in three dimensions. 
Another drawback is their limitation to benefit from the abundant 2D data,
since manually annotated images have no associated 3D information.

\rev{A common approach to directly use 2D data during training is to first learn a 2D
pose estimator, than} lift 3D poses from 2D estimations~\cite{Lee_2018_ECCV, Hossain_2018_ECCV,
Yang_2018_CVPR, Tome_2017_CVPR, Martinez_2017, Chen_2017_CVPR}.  However,
lifting 3D from 2D points only is an ill-defined problem \rev{since no visual
cues are available}, frequently resulting in ambiguity and, consequently,
limited precision.
Other methods assume that the absolute location of the root joint is provided
during inference~\cite{Zhou_2017_ICCV, Luvizon_2018_CVPR}, \rev{so the inverse
projection from pixels to millimeters can be performed}.
In our approach, this assumptions is not made, since we estimate
the 3D pose in absolute coordinates. The only additional information we need
are the intrinsic parameters for monocular prediction, which is
often given by the manufacturer or could be \revb{estimated by} standard tools.
\revb{In addition and differently from \cite{Luvizon_2018_CVPR}, our approach allows combining predictions from multiple views of the scene, resulting in more precise estimations}.

Contrarily to the previous work, we are able to train our method
simultaneously with 3D and 2D annotated data in an effective way, since one
part of our prediction is performed in the image plane and completely
independent from 3D information. Moreover, estimating the first two coordinates
in pixels in the image plane is a better defined problem than estimating
floating 3D positions directly in millimeters. These advantages translate into
higher accuracy for our method.

\subsection{Monocular absolute 3D human pose estimation}

\rev{
Contrarily to relative estimation, in absolute
pose prediction the 3D coordinates of the human body are predicted with respect
to the camera \revb{or in the view frustum space}.
A simple approach is to infer the distance to the camera considering
a normalized or constant body size~\revb{\cite{Zhou_2016_CVPR, VNect_SIGGRAPH2017}, 
which is an information that may not be available and difficult to be estimated~\cite{Gunel_2019_ICCV}}.
Inspired by the many works on depth estimation, \revb{Nie et al.}~\cite{nie2017monocular}
predict the depth of body joints individually.  The drawback of this
method is that it suffers to capture the human body structure, since errors in
the estimated depth for individual joints can degenerate the final pose.

More recently, multi-person absolute pose estimation methods were proposed~\revb{\cite{Moon_2019_ICCV, XNect_SIGGRAPH2020}}. \revb{In \cite{Moon_2019_ICCV}}, the absolute distance
from the person to the camera \revb{is predicted} based on the area of the cropped 2D bounding box.
However, it is known from the literature on absolute depth
estimation~\cite{Dijk_2019_ICCV, NIPS2016_6489} that not only the size of
objects are important, but also \revb{their positions} in the image is an
informative cue to predict its depth. For example, a person in the bottom of an
image is more likely to be closer to the camera than a person on the top of the
same image. \revb{Differently, in \cite{XNect_SIGGRAPH2020}, the authors optimized
the person absolute distance based on the initial bone lengths, estimated from the first 10 frames of a video sequence, and on the re-projection of the 3D pose into the 2D body joint locations. Besides this approach relies on video sequences, it also requires the camera parameters.}
}

In our approach, we combine three different information to predict the distance
of the root joint \revb{w.r.t. the camera position}: the size of the bounding box (including its ratio), the \revb{target}
position in the image, and deep convolutional features that provide additional
visual cues.

\subsection{\revb{Multi-view pose estimation and camera calibration}}

\rev{
For the challenging cases of occlusion or clutter background, multiple views
can be decisive to disambiguate uncertain positions of body joints (see
Fig.~\ref{fig:intro}).
To handle this,} several works have proposed multi-view solutions for 3D
human pose estimation~\cite{belagiannis20143d, amin2013multi, burenius20133d,
Belagiannis_PAMI_2016, Hofmann2012}, mostly exploring the classical concept of
pictorial structures from multi-view images.
Deep neural networks have been used to estimate relative
3D poses from a set of 2D predictions from different
views~\cite{rhodin2018learning, pavlakos2017harvesting, NUNEZ2019335}.
As an example, Pavlakos \etal~\cite{pavlakos2017harvesting} proposed to
collect 3D poses from 2D multi-view image, which are used to learn
a second model to perform 3D estimations.
Since these methods estimate 3D from multiple 2D images, they often require
both intrinsic and extrinsic parameters.

\revb{
In order to estimate the full camera calibration parameters, Micusik and
Pajdla~\cite{micusik2010simultaneous} proposed to use a human body seen
at different positions in the image. The main limitation of this approach
is the fact that it assumes that all poses as nearly vertical and parallel
to each other. Considering multiple views of the same person,
Rhodin \etal~\cite{rhodin2018learning} propose to estimate 3D poses from each
individual view and to estimate the extrinsic camera calibration, assuming
that the intrinsic parameters are provided as input.
More recently, Iskakov \etal~\cite{iskakov2019learnable} proposed a learnable
triangulation of 3D poses, considering multiple fully calibrated views
during training. Despite the impressive results achieved in this work,
the network model is training considering a pre-defined camera positioning,
which could result in a strong overfiting in the experimental setup.
Differently, our model is trained without priors about the camera positions and
the proposed multi-view optimization algorithm is not directly tied to a specific camera setup.
}

From the recent literature, we can notice that current multi-view approaches
are still completely dependent on the camera intrinsic parameters and often
require a complete calibration setup, which can be prohibitive in some
circumstances. Available methods are also limited to the inference of 3D from
multiple 2D predictions, requiring multi-view datasets for training.
Alternatively, we propose to predict absolute 3D poses from each individual
view, which has two important advantages over previous methods. \textit{First}, it
allows us to easily combine predictions from multiple calibrated cameras, while
requiring a single monocular training procedure. \textit{Second}, we are able to
estimate camera calibration, both intrinsic and extrinsic, from multi-view
images, by a consensus-based optimization without retraining the model.  The
strength of our approach is evidenced by its strong results, even when
considering unknown and uncalibrated cameras.

\section{Proposed method for 3D human pose estimation}
\label{sec:method-3dpose}

One of the goals of our method is to predict 3D human poses in absolute coordinates
with respect to the camera position.  For this, we believe that the most
effective approach is to predict each body joint in image pixel coordinates and
in absolute depth, orthogonal to the image plane, in millimeters. Then, the
predicted
pixel coordinates and depth can be projected to the world coordinates, considering a
pinhole camera model.

We further split the problem as relative 3D pose estimation and absolute depth
estimation. The motivation for this comes from the idea that a well cropped
bounding box around the person is better for predicting its pose than the full
image frame, since a better resolution could be attained and the person scale
is automatically handled by the image crop, \revb{although small variations in the bounding boxes during training result in better robustness}. \rev{Additionally, by providing a
separated loss on relative depth for each joint helps the network to learn
the human body structure, which would be more difficult to learn directly from
absolute coordinates due to position shift.
}

Recent works on depth estimation have demonstrated that neural networks rely on
both \textit{pictorial cues} and \textit{geometry} information to predict
depth \cite{Dijk_2019_ICCV}.
For the specific problem of 3D human pose estimation, the structure of the
human body is also an important \textit{domain knowledge} to be explored.
Considering our motivations and the exposed challenges, we propose to predict
3D poses relative to a cropped region \rev{centered at the person}, \rev{which
eases the network to encode the human body structure,} and absolute
depth from combined local pictorial cues and global position and size of
the cropped region.

\begin{figure}[ht]
  \centering
  \includegraphics[width=0.48\textwidth]{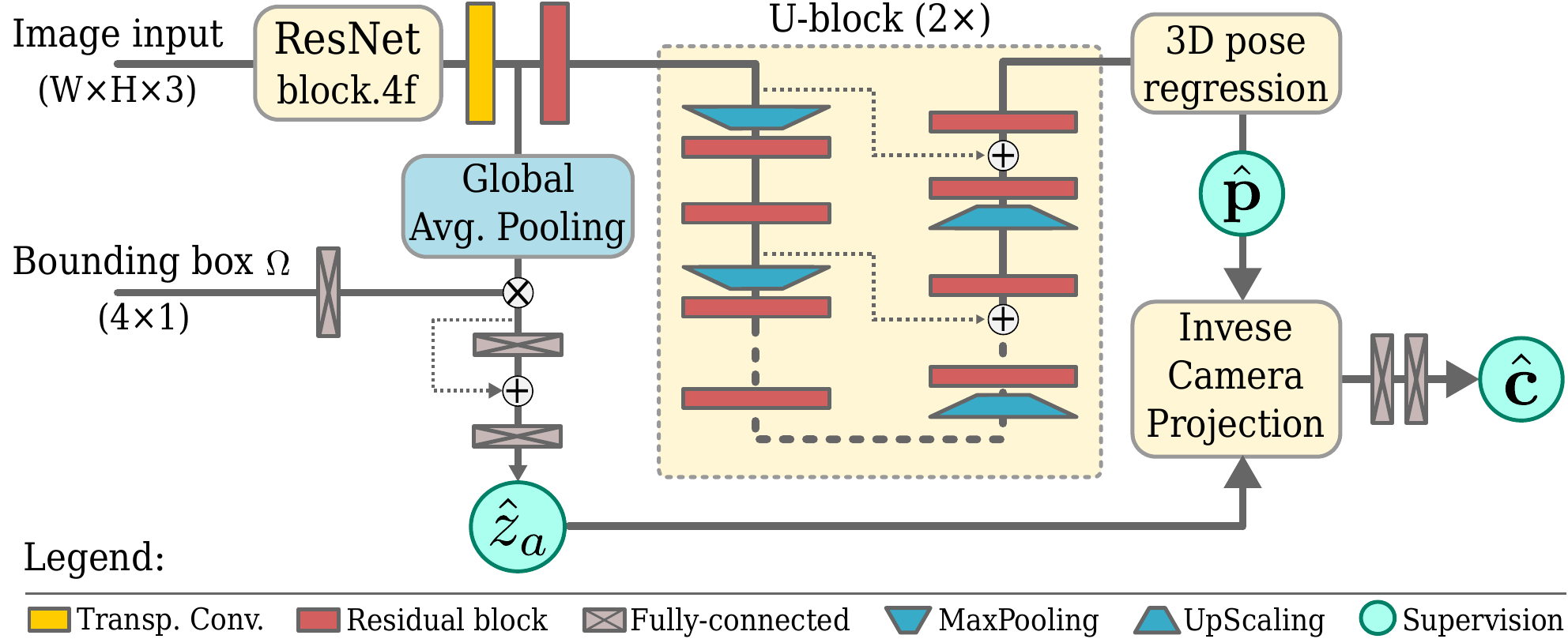}
  \caption{Proposed ResNet-U architecture. \revb{A given input image and
  the corresponding} bounding box parameters are fed into a neural network that predicts
  absolute depth $\eza{}$, human pose $\ep{}$, and confidence scores $\ec{}$.
  }
  \label{fig:net-arch}
\end{figure}

Specifically, given an image $\mathbf{I}\in\mathbb{R}^{W\times{H}\times{3}}$
and a person bounding box region $\Omega\in\mathbb{R}^{4\times1}$, we define
the problem as learning a function $\mathcal{F}$:
$\{\mathbf{I},\Omega\}\xrightarrow[]{\mathcal{F}}~\{\ep{},\ec{},\eza{}\}$,
where $\ep{}\in\mathbb{R}^{3\times{\nj{}}}$ is the estimated relative pose,
composed of $\nj{}$ body joints in the format $(\eui{},\evi{},\edi{})^T$, with
$i=\{1,\dots,\nj{}\}$, $\ec{}\in\mathbb{R}^{1\times\nj{}}$ contains the
\rev{body joint \textit{confidence score}}, which is an additional
information that represents \rev{a level of confidence for each predicted body
joint coordinates}, and $\eza{}\in\mathbb{R}_{\geq0}$ is the estimated absolute
depth for the person's root joint.
The person bounding box $\Omega$ is defined by its central position
$(x_{\Omega},y_{\Omega})$ and size $(w_{\Omega},h_{\Omega})$, and can be
obtained using a standard person detector~\cite{Redmon_2017_CVPR}.
The parametrized regression function $\mathcal{F}$
is implemented as a deep convolutional neural network (CNN), detailed as
follows.

\subsection{Network architecture}

U-Nets are widely used for human pose estimation due to their multi-scale
processing capability \cite{Newell_ECCV_2016}, while classic residual networks \cite{he2016deep}
are often preferable to produce CNN features. Since we want precise pose
predictions and informative visual features for absolute depth estimation,
\revb{our network combines} a ResNet-50 model cut at block 4f as
backbone and U-blocks, as shown in Fig.~\ref{fig:net-arch}. This
architecture is called ResNet-U and, \rev{in addition to a few fully connected
layers to regress the absolute depth $\eza{}$ and the confidence
scores $\ec{}$,} implements the function $\mathcal{F}$.
\rev{The details about each part of our method is discussed as follows.}

\subsection{3D human pose estimation}

As previously stated, we first want to estimate the 3D human pose
relative to the cropped bounding box. To this end, we first predict the pixel
coordinates $(\eui{},\evi{})$ in the image plane, given the information about
the cropped image in $\Omega$. Since it is difficult to predict the absolute depth
from an arbitrarily cropped region, at this stage we predict the relative depth
of each body joint with respect to the location of the person.
Therefore, the human pose estimation problem can be naturally split into
two parts: relative pose estimation and absolute body joints depth estimation,
as detailed next.

\subsubsection{Relative human pose estimation}

For the pose prediction in the image plane $(u, v)$, we use the
soft-argmax operation~\cite{LUVIZON201915, Yi_2016}, which induces
the U-Nets to generate one probability distribution per body joint.
This probability distribution
is defined as a feature map $\h{}_i\in\mathbb{R}^{w_f\times{h_f}}$
(positive and unitary sum) for the $i$th body joint.
\rev{
The third dimension of the pose $\ep{}$ is composed of the depth per
body joint, with respect to the location of the person.
}
This prediction could be
integrated in the soft-argmax by extending the feature map $\h{}$ to a
volumetric representation~\cite{Sun_2018_ECCV}.  However, depending on the
resolution and the number of body joints, this approach can be costly.
Instead, we predict a normalized \revb{depth map
$\dd{}\in\mathbb{R}^{w_f\times{h_f}}$ per body joint, corresponding to an
interval of 2 meters, which is a common reference size in other methods~\cite{Pavlakos_2017_CVPR}.}
By restricting the estimated depth to this range, we ensure that the bounding
box prediction is well defined inside an small region,
corresponding to the enclosure of an average person.
The regressed depth inside the bounding box is defined by:
\revb{
\begin{equation}
  \edi{} = \lambda\Big(\sum_{\y{}=(1,1)}^{(w_f, h_f)}d_i(\y{})h_i(\y{})\Big),
    \label{eq:depth}
\end{equation}
where $\lambda$ is a normalization function, $d_i(\y{})$ and $h_i(\y{})$ are the values of the depth and the joint probability distribution for the $i$th joint at pixel $\y{}$.}
Note that in Equation~\ref{eq:depth} \revb{the regions in the depth maps are pooled 
accordingly} to the high probability locations of the body joints.

\subsubsection{Absolute depth estimation}

Once we have estimated the body joint coordinates in pixels and the depth with
respect to the location of the person, we then predict the absolute depth of
the person with respect to the camera. For this, we use two
\rev{complementary} sources of 
information: the bounding box information (position and size), and deep visual features.
The position and size of the bounding box provide a rough global information
about the scale and position of the person in the image.
Additionally, the visual features extracted from the bounding box region by
means of ResNet features provide informative visual cues that are used to
refine the absolute person distance estimation.

Both extracted features are then fed to a fully connected network with 256
neurons at each level and a single neuron as output,
represented by $\alpha_z$, which is activated by a smoothed sigmoid
function, defined as:
\begin{equation}
  \eza{} = \rho\frac{1}{1 + e^{\beta\alpha_z}},
    \label{eq:za}
\end{equation}
where $\rho$ is the maximum depth, set as 10 meters in our experiments, \revb{and $\beta$ is a smoothing factor, set to 0.5 in our experiments}. The
output $\eza{}$ is then supervised with the absolute depth
of the root joint $\mathit{z}_a$.
This process is illustrates in Fig.~\ref{fig:net-arch} in the bottom left part.
We demonstrated in our experiments that the two different types of information,
visual and bounding box information, are complementary for the task of
absolute depth prediction.

\subsubsection{Absolute 3D human pose reconstruction}

In order to accomplish our objective of estimating the absolute 3D pose
represented as $\ex{} = (\exi{}, \eyi{}, \ezi{})^T$, we
combine the estimated pose in the bounding box with the predicted absolute
depth. Considering $(\eui{}, \evi{}, \edi{})^T$ as the first and $\eza{}$
as the last, the absolute $z$ coordinate for each body joint is defined
by $\ezi{} = \edi{} + \eza{}$.
Note that $\ezi{}$ is the absolute distance in millimeters from each body joint
to the camera and it results from the combination of two distinct predictions,
which are individually supervised.
The other two absolute coordinates required to build the final absolute 3D
pose, $\exi{}$ and $\eyi{}$, can then be computed using the pinhole camera model by the following equation:
\begin{equation}
  \begin{bmatrix}
    \exi{} \\
    \eyi{}
  \end{bmatrix}%
  =%
  \begin{bmatrix}
    1/\fx{} & 0 & -\cx{}/\fx{}\ \\
    0 & 1/\fy{} & -\cy{}/\fy{} \\
  \end{bmatrix}%
  \begin{bmatrix}
    \eui{} \\
    \evi{} \\
    1
  \end{bmatrix}%
  \ezi{},
  \label{eq:xy-prediction}
\end{equation}
where $f$ and $C$ are the camera focal length and the camera center, both in
pixels, considering the $x$ and $y$ axis.
Note that these parameters are camera intrinsics and can be easily obtained.
The camera focal length is often given by the manufacturer and the center
of the image frequently corresponds to the image center in pixels or, even,
both values could be estimated with standard tools.
Nevertheless, \revb{in what follows we present} a method to estimate the
camera parameters, both intrinsic and extrinsic, without any prior information, directly
from the predictions of our method, considering a multi-view scenario
with uncalibrated cameras.

\section{Consensus-based optimization}
\label{sec:consensus-optimization}

One of the main advantages of estimating absolute instead of relative 3D poses
is the possibility to project the predictions from one camera to another,
simply by applying a rotation and a translation.  This advantage has important
consequences in multi-view. For example, when the camera calibration is known,
the predictions of different monocular cameras can be combined with respect to a common
reference, resulting in more precise predictions.
For the cases where no
information is known about the camera calibration, we propose a consensus-based
algorithm that can be applied to estimate both intrinsic and extrinsic
parameters, resulting in a completely uncalibrated multi-view approach. This
algorithm is explained as follows.

Let us define the predictions of the proposed method from two distinct cameras as:
$\ep{}_{c1} = (\eui{}, \evi{}, \ezi{})_{c1}^T$ and
$\ep{}_{c2} = (\eui{}, \evi{}, \ezi{})_{c2}^T$, and
their poses in absolute camera coordinates:
$\ex{}_{c1} = (\exi{}, \eyi{}, \ezi{})_{c1}^T$ and
$\ex{}_{c2} = (\exi{}, \eyi{}, \ezi{})_{c2}^T$, respectively for cameras 1 and 2.
Then, we define the projection of $\ex{}_{c2}$ into camera 1 as:
\begin{equation}
  \ex{}_{c2\rightarrow{c1}} = \textbf{R}_{2,1}(\ex{}_{c2} - \textbf{T}_{2,1}),
  \label{eq:camera-ext}
\end{equation}
where $\textbf{R}_{2,1}\in\mathbb{R}^{3\times3}$ and
$\textbf{T}_{2,1}\in\mathbb{R}^{3\times1}$ are a rotation
matrix and a translation vector from camera 2 to camera 1.
Our goal is to minimize the projection error from camera 2 to camera 1 (and
vice-versa) by optimizing a set of camera parameters:\\
$\textbf{K}_{2,1}\in\{\fx{}_1, \fy{}_1, \cx{}_1, \cy{}_1, \fx{}_2, \fy{}_2,
\cx{}_2, \cy{}_2, \textbf{R}_{2,1}, \textbf{T}_{2,1}\}$, which includes the
intrinsics from both cameras and the extrinsic parameters between them.
Specifically, let us define the optimization problem as:
\begin{equation}
  \textbf{K}^{\ast}_{2,1}=%
    \argmin_{\textbf{K}_{2,1}}\|\ex{}_{c1}-\ex{}_{c2\rightarrow{c1}}\|_F^2.
  \label{eq:optimization-camera}
\end{equation}
We find a solution for Equation~\ref{eq:optimization-camera} by using an optimization approach that sequentially considers the individual variables by alternating gradient with
steepest descent. This process is detailed as follows.

\subsection{Camera parameters optimization}

In order to obtain the translation vector from camera 2 to camera 1 that
minimizes Equation~\ref{eq:optimization-camera}, we define:
\begin{equation}
  \textbf{T}^\ast_{2,1}=%
    \argmin_{\textbf{T}_{2,1}}%
    \|\ex{}_{c1}-\textbf{R}_{2,1}(\ex{}_{c2}-\textbf{T}_{2,1})\|_F^2.
\end{equation}
By replacing the squared Frobenius norm $\|\mathbf{M}\|_F^2$ by
$\Tr{}(\mathbf{M}^T\mathbf{M})$ and by optimizing for
$\nabla_{\textbf{T}_{2,1}}=0$, we obtain:
\begin{equation}
  -\ex{}_{c2}+\textbf{T}^{\prime}_{2,1}+\textbf{R}_{2,1}^T\ex{}_{c1}=0.
  \label{eq:deriv-t}
\end{equation}
In Equation~\ref{eq:deriv-t}, we are considering $\textbf{T}^{\prime}_{2,1}$ with shape
$\mathbb{R}^{3\times{N}}$ to simplify notation. Therefore, for a set of
$N$ points, we assume a single translation as the average of the individual
solutions for all points, which results in:
\begin{equation}
  \textbf{T}^\ast_{2,1} = \frac{1}{N}\sum_{i=1}^{N}(\ex{}_{c2}-\textbf{R}_{2,1}^T\ex{}_{c1}).
  \label{eq:optimal-t2}
\end{equation}
Once the poses from both cameras are aligned, the rotation matrix $\textbf{R}_{2,1}$ can be
updated with rigid Procrustes alignment.

For the camera intrinsic parameters, we can also re-write
Equation~\ref{eq:optimization-camera} for the
focal length and camera center only, considering the camera projection
(Equation~\ref{eq:xy-prediction}), resulting in:
\begin{equation}
  \fx{}^\ast_1,\cx{}^\ast_1=\argmin_{\fx{}_1,\cx{}_1}%
    \Big\|\frac{(\eui{}_1-\cx{}_1)}{\fx{}_1}\ezi{}_1-\excfrom{}\Big\|_F^2.
    \label{eq:camera-intrisic-eq}
\end{equation}
To obtain the focal length and the camera center that minimize Equation~\ref{eq:camera-intrisic-eq},
we can re-write the individual optimizations as:
\begin{equation}
  \fpx{}^\ast_1=\argmin_{\fpx{}_1}%
    \|\fpx{}_1\mathbf{A}_{x1}-\ex{}^x_{c2\rightarrow{c1}}\|_F^2,
    \label{eq:opt-focal-only}
\end{equation}
\begin{equation}
  \cx{}^\ast_1=\argmin_{\cx{}_1}%
  \|(\eui{}_1-\cx{}_1)\mathbf{B}_{x1}-\excfrom{}\|_F^2,
  \label{eq:opt-center-only}
\end{equation}
where $\fpx{}_{1}=1/\fx{}_{1}$,
$\mathbf{A}_{x1}=(\eui{}_1-\cx{}_1)\ezi{}_1$, and
$\mathbf{B}_{x1}=\ezi{}_{c1}/\fx{}_1$.
By solving the Equations~\ref{eq:opt-focal-only} and \ref{eq:opt-center-only}
respectively for $\nabla_{\fpx{}_{1}}=0$ and $\nabla_{\cx{}_{1}}=0$, we finally obtain:
\begin{equation}
  \fx{}^\ast_1 = \frac{1}{\excfrom{}\textbf{A}^T_{x1}(\textbf{A}_{x1}\textbf{A}^T_{x1})^{-1}},
  \label{eq:optimal-f}
\end{equation}
\begin{equation}
  \cx{}^\ast_1 = (\eui{}_{c1}\textbf{B}_{x1} - \excfrom{})\textbf{B}^T_{x1}(\textbf{B}_{x1}\textbf{B}^T_{x1})^{-1}.
  \label{eq:optimal-c}
\end{equation}
Note that, for Equations~\ref{eq:optimal-f} and \ref{eq:optimal-c},
the intrinsic parameters for $\hat{y}$ follow a similar form, replacing
the $\hat{x}$ components by $\hat{y}$ and $\eui{}$ by $\evi{}$.
For the intrinsics from camera 2, the same equations are used,
except by swapping the camera indexes in each variable.
Additionally, the reverse projection
($\textbf{R}_{1,2}$ and $\textbf{T}_{1,2}$), from camera 1 to camera 2, is given by
isolating $\ex{}_{c2}$ from Equation~\ref{eq:camera-ext}.

In the video case, given two static cameras, we can use a sequence of frames
to estimate the camera calibration, from where we can obtain more points than
from a single frame and from a single pose.
Finally, we can solve the global optimization problem by alternating
the optimization of camera extrinsic and intrinsic parameters.
This process is detailed in Algorithm~\ref{alg:camera-params}.
\revb{Note that we initialize the camera rotation $\textbf{R}$ with the identity ($\textit{I}_3$)}.

\begin{algorithm}[!htb]
  \caption{Camera parameters optimization.}
  \label{alg:camera-params}
  \begin{algorithmic}[1]
    \Require $\ep{}_{c1}, \ep{}_{c2}$, $MaxIter$
    \State Initialize $\fx{}_1, \fy{}_1, \cx{}_1, \cy{}_1, \fx{}_2, \fy{}_2, \cx{}_2, \cy{}_2$
    \State Compute $\ex{}_{c1}$ and $\ex{}_{c2}$ from Equation~\ref{eq:xy-prediction}
    \State Initialize $\textbf{T}_{2,1}$ using Equation~\ref{eq:optimal-t2} (assume $\textbf{R}_{2,1}=\mathbf{\textit{I}}_3$)
    \State $Iter \gets 0$
    \Repeat
      \State Update $\textbf{R}_{2,1}$ using rigid Procrustes alignment
      \State Update $\textbf{T}_{2,1}$ using Equation~\ref{eq:optimal-t2}
      \If {$mod(Iter, 4) = 0$}
        \State Update $\fx{}_1, \fy{}_1$ using Equation~\ref{eq:optimal-f}
      \ElsIf {$mod(Iter, 4) = 1$}
        \State Update $\fx{}_2, \fy{}_2$ using Equation~\ref{eq:optimal-f}
      \ElsIf {$mod(Iter, 4) = 2$}
        \State Update $\cx{}_1, \cy{}_1$ using Equation~\ref{eq:optimal-c}
      \ElsIf {$mod(Iter, 4) = 3$}
        \State Update $\cx{}_2, \cy{}_2$ using Equation~\ref{eq:optimal-c}
      \EndIf
      \State Update $\ex{}_{c1}$ and $\ex{}_{c2}$ from Equation~\ref{eq:xy-prediction}
      \State $Iter\gets{Iter+1}$
    \Until {$(Iter < MaxIter)$}
    \State \textbf{return} $\fx{}_1, \fy{}_1, \cx{}_1, \cy{}_1, \fx{}_2, \fy{}_2, \cx{}_2, \cy{}_2, \textbf{R}_{2,1}, \textbf{T}_{2,1}$
  \end{algorithmic}
\end{algorithm}

\subsection{Body joint confidence scores}

\rev{
Since the proposed consensus-based optimization algorithm relies on estimated
poses, it can be affect by the precision of predicted joint positions. Despite
the average error of our method being very low compared to previous approaches, 
we also propose a confidence score that indicates whether the network
is ``confident'' or not for each predicted body joint. This score varies from
0 to 1, and is implemented by a DNN that takes estimated poses as input (see
Fig.\ref{fig:net-arch} - bottom right) \revb{and is trained by comparing predictions to the pose ground truth}. The ground truth for the $i$th joint is
defined as follows:
\begin{equation}
  \mathbf{c}_i = \frac{1}{1 + e^{(\overline{d} - d_i)/\sigma_d}},
  \label{eq:confidence}
\end{equation} where $d_i$ is the distance error between the predicted and
ground truth joint position, $\overline{d}$ is the average prediction error, and
$\sigma_d$ is the error standard deviation. By estimating
Equation~\ref{eq:confidence}, we can remove predicted joints with error higher
than the average simply by discarding points with $\ec{} < 0.5$ \revb{(binary decision)}.
The predicted confidence score $\ec{}$ is useful in
Algorithm~\ref{alg:camera-params}, \revb{providing a way to filter wrong predictions.}
In addition, we also take into account the confidence scores when predicting
poses in multi-view scenario by weighting each body joint from each view by its
corresponding predicted confidence score.
}

\section{Experiments}
\label{sec:experiments}

In this section, we present the results of our method on two well known
datasets, as well as a sequence of ablation studies to provide  insights about
our approach.

\subsection{Datasets}

\textbf{Human3.6M}~\cite{h36m_pami} is a large-scale dataset with 3D human
poses collected by a motion capture system (MoCap) and RGB images captured by 4
synchronized cameras.  A total of 15 activities are performed by 11 actors, 5
females and 6 males, resulting in 3.6 million images.  Poses are
composed of 23 body joints, from which 17 are used for evaluation \revb{as in the previous work~\cite{Pavlakos_2017_CVPR, Zanfir_2018_CVPR}}.

\noindent
\textbf{MPI-INF-3DHP}~\cite{Mehta_2017_3DV} is a dataset for 3D human
pose estimation captured with a marker-less MoCap system, which allows outdoor
video recording, \eg, TS5 and TS6 from testing. A total of 8 activities are
performed by 8 different actors in two distinct sequences.  Human poses are
composed of 28 body joints, from which 17 are used for evaluation.
The activities involve complex exercising poses, which makes this dataset
more challenging than Human3.6M. However, the precision of marker-less
motion capture is visually less precise than ground truth poses
from~\cite{h36m_pami}.  Despite having a training set captured by 8 different
cameras, test samples are captured by a single monocular camera.

\noindent
\revb{
\textbf{PennAction}~\cite{Zhang_PennAction} is a dataset composed by 2,326
videos in the wild with annotated 2D poses of people performing 15 different
actions. This dataset does not provide 3D pose annotations, but it is usefull to
access the generability of our method in a qualitative evaluation, since the
images are very challenging for pose estimation.
}

\noindent
\revb{
\textbf{KTH Multiview Football Dataset II}~\cite{kazemi2013multi} consists of images from football players with ground truth 2D and 3D poses centered in the root joint. Partial camera parameters are given for projecting the 3D poses into the three different views, however, explicit intrinsic and extrinsic parameters are not available. This dataset is challenging since the camera setup is very different from the training scenario on both Human3.6M and MPI-INF-3DHP. Therefore, we used KTH for zero-shot evaluation.
}

\subsection{Evaluation protocols and metrics}

Three evaluation protocols are widely used for Human-\\
3.6M. In \textit{protocol
1}, six subjects are used for training and only one is used for evaluation.
Since this protocol uses a Procrustes alignment between prediction and ground
truth, we do not consider it in our work.
In \textit{protocol 2}, five subjects (S1, S5, S6, S7, S8) are dedicated for
training and S9 and S11 for evaluation, and evaluation videos are
sub-sampled every 64th frames.
The third protocol is the official test set (S2, S3, S4), of which ground truth
poses are withheld by the authors and evaluation is performed over all test
frames (almost 1 million images) through a server. In our experiments,
we report our scores in the most challenging official test set. Additionally,
we consider \textit{protocol 2} for the ablation studies and for comparison
with multi-view approaches.

The standard metric for Human3.6M is the \textit{mean per joint position error}
(MPJPE), which measures the average joint error after centering both
predictions and ground truth poses to the origin.
\rev{
  We also evaluated our method considering the mean of the root joint position
  error (MRPE)~\cite{Moon_2019_ICCV}, which measures the average error related
  to the absolute pose estimation. This metric is considered only for
  validation, since the server does not support this protocol.
}

For MPI-INF-3DHP, evaluation is performed on a test set composed of
6 videos/subjects, of which 2 are recorded in outdoor scenes, resulting in
almost 25K frames.
The authors of~\cite{Mehta_2017_3DV} proposed three evaluation metrics: the mean
per joint position error, in millimeters,
the 3D Percentage of Correct Keypoints (PCK), and the Area Under the Curve (AUC)
for different thresholds on PCK. The standard threshold for PCK is 150mm~\revb{\cite{Mehta_2017_3DV}, which corresponds nearly to half of the head size}.
\revb{Differently from previous work~\cite{Mehta_2017_3DV, Kocabas_2019_CVPR, Zhou_2017_ICCV}}, we use the real 3D poses to compute the error
instead of the normalized 3D poses, since the last is not compatible with a constant
camera projection. Since evaluation is performed on monocular images, we use
the available intrinsic camera parameters to recover absolute poses in
millimeters.
\revb{Finally, we also evaluated our method on KTH considering the PCP metric from~\cite{burenius_3d_2013}.}

\subsection{Implementation details}

During training, we use the elastic net loss (L1+L2)~\cite{Zou05regularizationand} for both absolute
$z$ and relative 3D pose predictions, respectively defined by:
\begin{equation}
  \mathcal{L}_z = \frac{1}{N_s}\sum_{i=1}^{N_s}%
  \|z_{ai}-\eza{}_i\|_1+\|z_{ai}-\eza{}_i\|_2^2, and
  \label{eq:loss-az}
\end{equation}
\begin{equation}
  \mathcal{L}_p = \frac{1}{N_s}\sum_{i=1}^{N_s}%
  \|\mathbf{p}_{i}-\ep{}_i\|_1+\|\mathbf{p}_{i}-\ep{}_i\|_2^2,
  \label{eq:loss-pose}
\end{equation}
where $z_{ai}$ and $\eza{}_i$ are the ground truth and the estimated absolute
$z$ values, and $\mathbf{p}_{i}$ and $\ep{}_i$ are the ground truth and the
estimated 3D poses. The final loss is then represented
by $\mathcal{L} = \mathcal{L}_z + \mathcal{L}_p$.

Once the first part of our network is trained, we compute the average prediction
error $\overline{d}$ on training, which is used to train the confidence score
network using the mean average error (MAE).
RMSprop and Adam are used for optimization, respectively for the first and
second training processes, starting with a learning rate of $0.001$ and decreased
by $0.2$ after 150K and 170K iterations. Batches of 24 images are used.
The full training process takes less then two days with a GTX 1080 Ti GPU.  We augmented
the training data with common techniques, such as random rotations
($\pm30^\circ$), re-scaling (from 0.7 to 1.3), horizontal flipping, color gains
(from 0.9 to 1.1), and artificial occlusions with rectangular black boxes.
\revb{We also added some randomness in the cropped bounding boxes, on both position and size, in order to make the model more robust against variations in human detection}.
Additionally, we augmented the training data in a 50/50 ratio with 2D images
from MPII~\cite{Andriluka_2014_CVPR}, which becomes an standard
data augmentation technique for 3D human pose estimation.

\begin{table*}[ht]
  \centering
  \caption{Comparison with results from related methods on Human3.6M
  \textit{test set} using MPJPE (millimeters error) evaluation. \revb{Training data: Human3.6M and MPII}.
  }
  \label{tab:result-h36m-testset}
  \footnotesize
  \begin{tabular}{@{}l|cccccccc@{}}
    \hline
    Methods & Directions & Discussion & Eating & Greeting & Phoning & Posing & Purchases & Sitting \\ \hline
    
    Ionescu \etal~\cite{h36m_pami} & 152 & 153 & 125 & 171 & 135 & 180 & 162 & 168 \\
    Popa \etal~\cite{Popa_CVPR_2017} & 60 & 56 & 68 & 64 & 78 & 67 & 68 & 106 \\
    Zanfir \etal~\cite{Zanfir_2018_CVPR} & 54 & 54 & 63 & 59 & 72 & 61 & 68 & 101 \\
    Zanfir \etal~\cite{NIPS2018_8061} & 49 & 47 & 51 & 52 & 60 & 56 & 56 & 82 \\
    Shi \etal~\cite{shi2018fbipose} & 51 & 50 & 54 & 54 & 62 & 57 & 54 & 72 \\ 
    \hline
    \textbf{Ours} monocular & {42} & {44} & {52} & {47} & {54} &  {48} & {49} &  {66} \\
    \textbf{Ours} multi-view est. calib. & {40} & {36} & {44} & {39} & {44} &  {42} & {41} &  {66} \\
    \textbf{Ours} multi-view GT calib. & \textbf{31} & \textbf{33} & \textbf{41} & \textbf{34} & \textbf{41} &  \textbf{37} & \textbf{37} &  \textbf{51} \\ \hline
    \hline
    Methods & Sit. Down & Smoking & Photo & Waiting & Walking & Walk.Dog & Walk.Pair & \multicolumn{1}{|c}{Average} \\ \hline
    
    Popa \etal~\cite{Popa_CVPR_2017} & 119 & 77 & 85 & 64 & 57 & 78 & 62 & \multicolumn{1}{|c}{73} \\
    Zanfir \etal~\cite{Zanfir_2018_CVPR} & 109 & 74 & 81 & 62 & 55 & 75 & 60 & \multicolumn{1}{|c}{69} \\
    Zanfir \etal~\cite{NIPS2018_8061} & 94 & 64 & 69 & 61 & 48 & 66 & 49 & \multicolumn{1}{|c}{60} \\ 
    Shi \etal~\cite{shi2018fbipose} & 76 & 62 & 65 & 59 & 49 & 61 & 54 & \multicolumn{1}{|c}{58} \\
    \hline 
    \textbf{Ours} monocular & {76} & {54} & {61} & {47} & {44} & {55} & {44} & \multicolumn{1}{|c}{{52}} \\
    \textbf{Ours} multi-view est. calib. & {70} & {46} & {49} & {43} & {34} & {46} & {34} & \multicolumn{1}{|c}{{45}} \\
    \textbf{Ours} multi-view GT calib. & \textbf{56} & \textbf{43} & \textbf{44} & \textbf{37} & \textbf{33} & \textbf{42} & \textbf{32} & \multicolumn{1}{|c}{\textbf{39}} \\\hline
  \end{tabular}
\end{table*}

\begin{table*}[ht]
  \centering
  \caption{Comparison with related multi-view methods on Human3.6M
  \textit{validation set}, protocol 2.  We report our scores in mm error
  (MPJPE), considering ground truth and estimated camera calibration.  Note
  that all previous methods use ground truth camera calibration. \revb{Training data: Human3.6M and MPII}.
  }
  \label{tab:result-h36m-validation}
  \footnotesize
  \begin{tabular}{@{}l|c|cccccccc@{}}
    \hline
    Methods &  \small{Cam. calib.} & Dir. & Discussion & Eating & Greeting & Phoning & Posing & Purchases & Sit. \\ \hline

    PVH-TSP~\cite{Trumble:BMVC:2017} & \small{GT} & 92.7 & 85.9 & 72.3 & 93.2 & 86.2 & 101.2 & 75.1 & 78.0 \\
    Trumble \etal~\cite{trumble2018deep} & \small{GT} & 41.7 & 43.2 & 52.9 & 70.0 & 64.9 & 83.0 & 57.3 & 63.5 \\
    Pavlakos \etal~\cite{pavlakos2017harvesting} & \small{GT} & 41.1 & 49.1 & 42.7 & 43.4 & 55.6 & 46.9 & 40.3 & 63.6 \\
    \revb{Tome \etal~\cite{tome2018rethinking}} & \small{GT} & 43.3 & 49.6 & 42.0 & 48.8 & 51.1 & 40.3 & 43.3 & 66.0 \\
    \revb{Kadkho. \etal~\cite{kadkhodamohammadi2021generalizable}} & \small{GT} & 39.4 & 46.9 & 41.0 & 42.7 & 53.6 & 41.4 & 50.0 & 59.9 \\
    \revb{Iskakov \etal~\cite{iskakov2019learnable}} & \small{GT} & \textbf{19.9} & \textbf{20.0} & \textbf{18.9} & \textbf{18.5} & \textbf{20.5} & \textbf{18.4} & \textbf{22.1} & \textbf{22.5} \\
    \hline
    \textbf{Ours} & \small{Estimated} & 59.3 & {40.7} & {38.7} & {39.1} & {41.7} & {39.5} & {40.6} & {64.1} \\
    \textbf{Ours} & \small{GT} & {31.0} & {33.7} & {33.8} & {33.4} & {38.6} & {32.2} & {36.3} & {48.2} \\ \hline
    \hline
    Methods & \small{Cam. calib.} & Sit. D. & Smoking & Photo & Waiting & Walking & Walk.Dog & Walk.Pair & \multicolumn{1}{|c}{Avg} \\ \hline

    PVH-TSP~\cite{Trumble:BMVC:2017} & \small{GT} & 83.5 & 94.8 & 85.8 & 82.0 & 114.6 & 94.9 & 79.7 & \multicolumn{1}{|c}{87.3} \\
    Trumble \etal~\cite{trumble2018deep} & \small{GT} & 61.0 & 95.0 & 70.0 & 62.3 & 66.2 & 53.7 & 52.4 & \multicolumn{1}{|c}{62.5} \\
    Pavlakos \etal~\cite{pavlakos2017harvesting} & \small{GT} & 97.5 & 119.9 & 52.1 & 42.6 & 51.9 & 41.7 & 39.3 & \multicolumn{1}{|c}{56.8} \\
    \revb{Tome \etal~\cite{tome2018rethinking}} & \small{GT} & 95.2 & 50.2 & 64.3 & 52.2 & 43.9 & 51.1 & 45.3 & \multicolumn{1}{|c}{52.8} \\
    \revb{Kadkho. \etal~\cite{kadkhodamohammadi2021generalizable}} & \small{GT} & 78.8 & 49.8 & 54.8 & 46.2 & 51.1 & 40.5 & 41.0 & \multicolumn{1}{|c}{49.1} \\
    \revb{Iskakov \etal~\cite{iskakov2019learnable}} & \small{GT} & \textbf{28.7} & \textbf{21.2} & \textbf{19.4} & \textbf{20.8} & \textbf{22.1} & \textbf{19.7} & \textbf{20.2} & \multicolumn{1}{|c}{\textbf{20.8}} \\
    \hline
    \textbf{Ours} & \small{Estimated} & {69.5} & {42.0} & {44.6} & {39.6} & {31.0} & {40.2} & {35.3} & \multicolumn{1}{|c}{{44.7}} \\
    \textbf{Ours} & \small{GT} & {51.5} & {39.2} & {38.8} & {32.4 } & {29.6} & {38.9} & {33.2} & \multicolumn{1}{|c}{{36.9}} \\
    \hline
  \end{tabular}
\end{table*}

\begin{table*}[htbp]
 \centering
  \caption{Results on MPI-INF-3DHP compared to the state of the art. \revb{Training data: MPI-INF-3DHP, Human3.6M, and MPII}}
  \label{tab:result-mpi3d}
  \small
\begin{tabular}{@{}l|ccccccc|ccc@{}}
\hline
\multirow{2}{*}{Method} & Stand & Exercise & Sit & Crouch & On the Floor & Sports & Misc. & \multicolumn{3}{c}{Total} \\
                        & PCK   & PCK      & PCK & PCK    & PCK          & PCK    & PCK   & PCK  & AUC & MPJPE  \\ \hline
  Rogez \etal~\cite{Rogez_CVPR_2017}\textsuperscript{$\star$} & 70.5 & 56.3 & 58.5 & 69.4 & 39.6 & 57.7 & 57.6 & 59.7 & 27.6 & 158.4 \\
  Zhou \etal~\cite{Zhou_2017_ICCV}\textsuperscript{$\star$} & 85.4 & 71.0 & 60.7 & 71.4 & 37.8 & 70.9 & 74.4 & 69.2 & 32.5 & 137.1 \\ 
  Mehta \etal~\cite{Mehta_2017_3DV}\textsuperscript{$\star$} & \textbf{86.6} & 75.3 & 74.8 & 73.7 & 52.2 & 82.1 & 77.5 & 75.7 & 39.3 & 117.6 \\
  Kocabas \etal~\cite{Kocabas_2019_CVPR}\textsuperscript{$\star$} & -- & -- & -- & -- & -- & -- & -- & 77.5 & -- & 108.99 \\
  \revb{Kolotouros \etal~\cite{Kolotouros_2019_ICCV}} & -- & -- & -- & -- & -- & -- & -- & 76.4 & 37.1 & \textbf{105.2}\\ \hline

\textbf{Ours} monocular & 83.8 &  \textbf{79.6} & \textbf{79.4} & \textbf{78.2} & \textbf{73.0} & \textbf{88.5} & \textbf{81.6} & \textbf{80.6} &  \textbf{42.1} & 112.1  \\ \hline
\end{tabular}\\
  \small\textsuperscript{$\star$} Methods using normalized 3D human poses for evaluation.
\end{table*}

\subsection{Comparison with the state of the art}

\noindent
\textbf{Human3.6M}. In Table~\ref{tab:result-h36m-testset}, we show our results on the test set
from Human3.6M.
We provide results of our method considering monocular predictions and
multi-view predictions, for estimated and ground truth camera calibration.  In
all the cases our method obtains state-of-the-art results by a fair merging,
reducing the prediction error by more than 10\% in monocular scenario.
In the multi-view setup, our method achieves 39mm error, reducing errors
by more than 32\% on average.
In the most challenging activity (Sitting Down), our method performs better
than all previous approaches \revb{reporting results in the official test set}.
These results demonstrate the effectiveness of our method, considering that
the test set from Human3.6M is very challenging and labels are withheld by the
authors.

For a fairer comparison, we also consider results only from multi-view
approaches in Table~\ref{tab:result-h36m-validation}. We present our scores
considering ground truth and estimated camera calibration, while all previous
methods use the available calibration from the dataset. Still, our method
obtains 36.9mm error, which \revb{is a strong results, specially considering that
the methods from \cite{iskakov2019learnable, tome2018rethinking} require
multi-view training with a known calibration setup,
while our network is trained with monocular images}.
In this comparison, we are not considering methods that make use of
the ground truth absolute position of the root joint, since in our method
we estimate this information.

\noindent
\textbf{MPI-INF-3DHP}. Our results on MPI-INF-3DHP are shown in Table~\ref{tab:result-mpi3d}.
We do not report results considering multiple views in this dataset,
since the testing samples were captured by a single camera.
Contrarily to what is more common in this dataset, we evaluated our method
using non-normalized 3D poses, otherwise it will not be possible to perform
the inverse camera projection.
Nevertheless, our method achieves results comparable to the state of the art,
even considering other methods using normalized 3D poses.

\begin{figure}[!h]
  \centering
  \includegraphics[height=2.23cm]{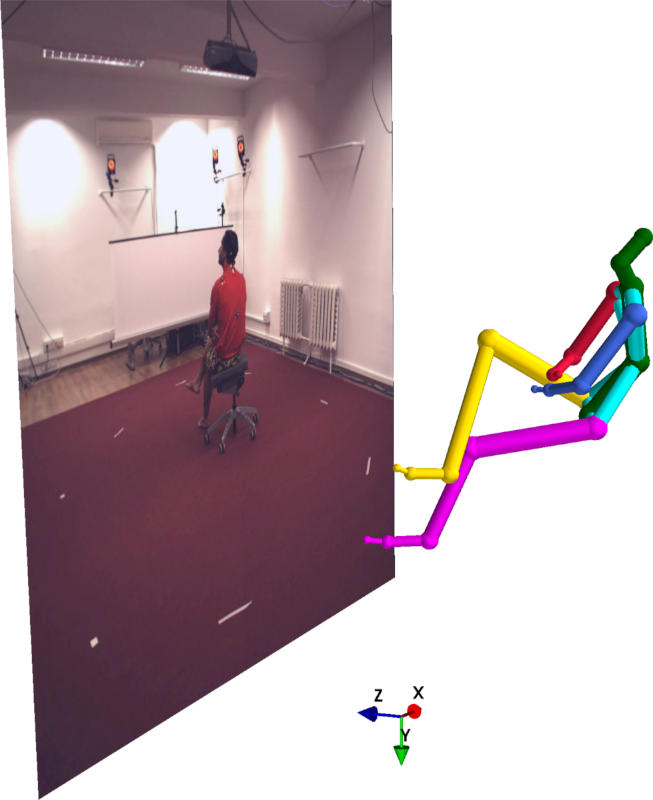}\hspace{0.05cm}
  \includegraphics[height=2.23cm]{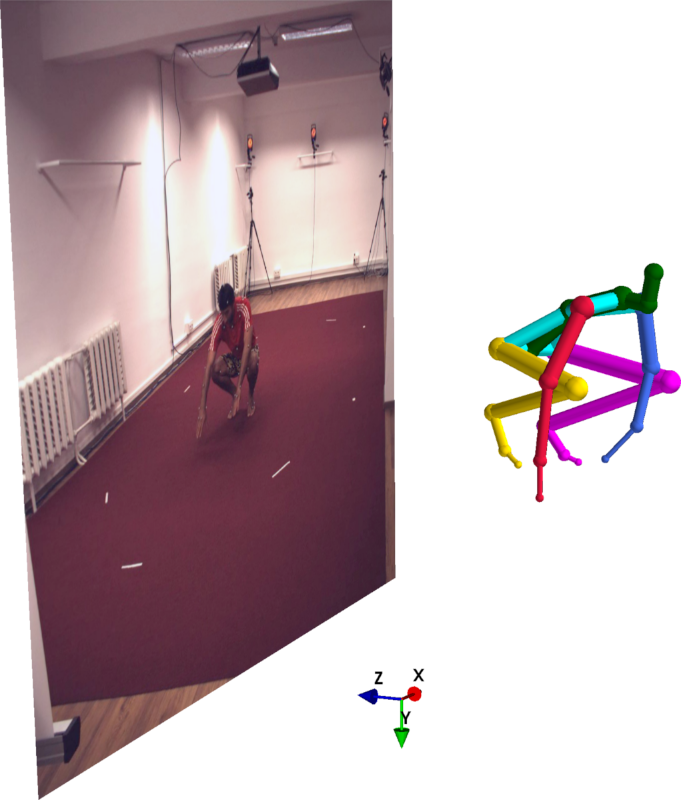}\hspace{0.05cm}
  \includegraphics[height=2.23cm]{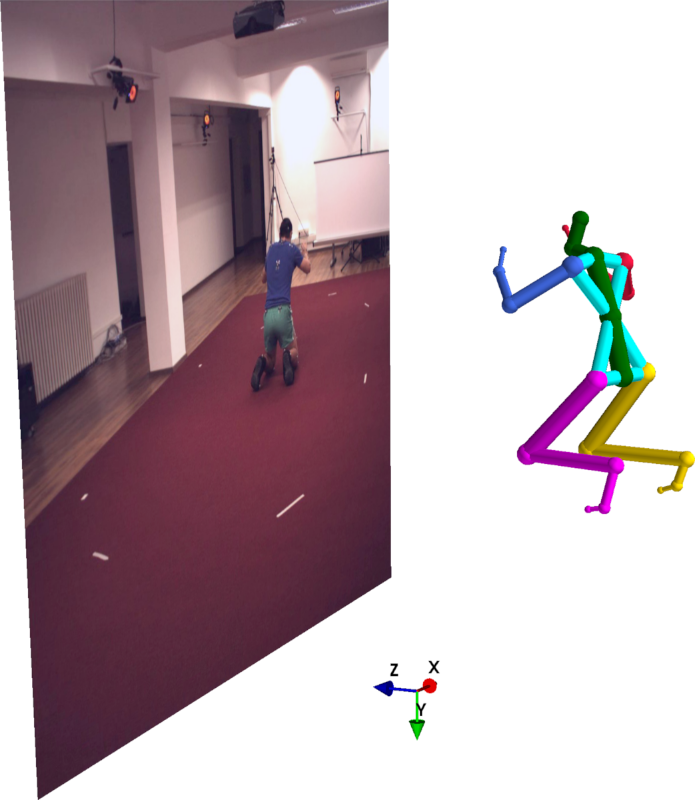}\hspace{0.05cm}
  \includegraphics[height=2.23cm]{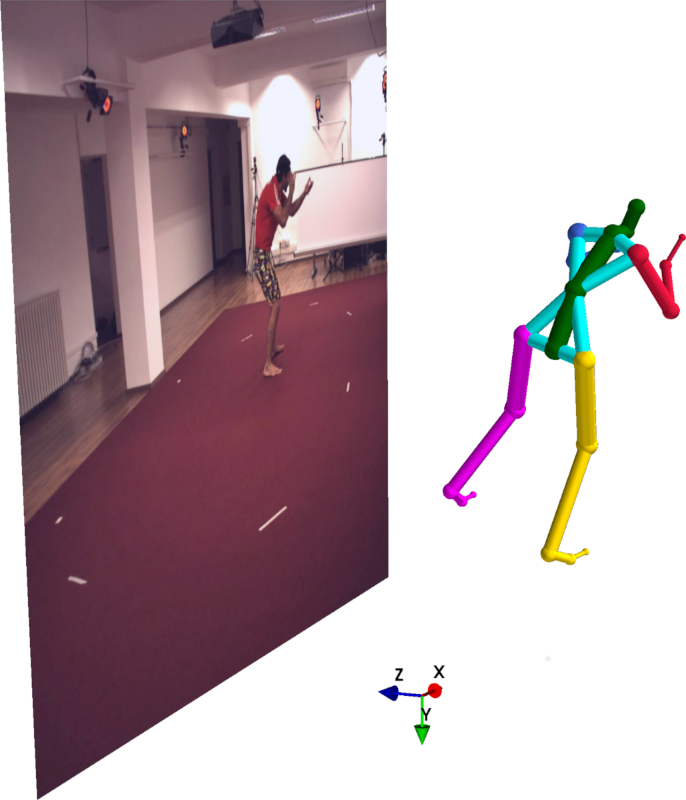}\\\vspace{0.05cm}
  \includegraphics[height=2.23cm]{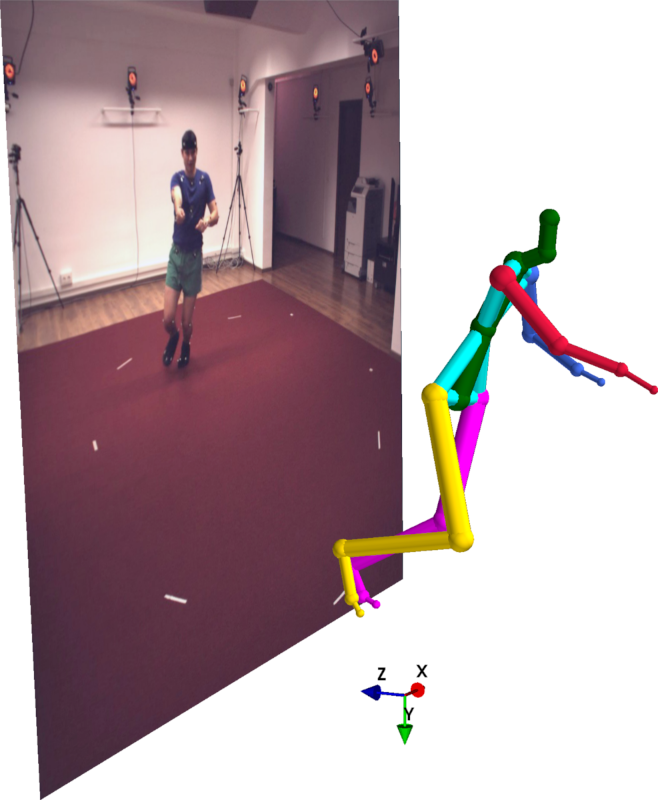}\hspace{0.05cm}
  \includegraphics[height=2.23cm]{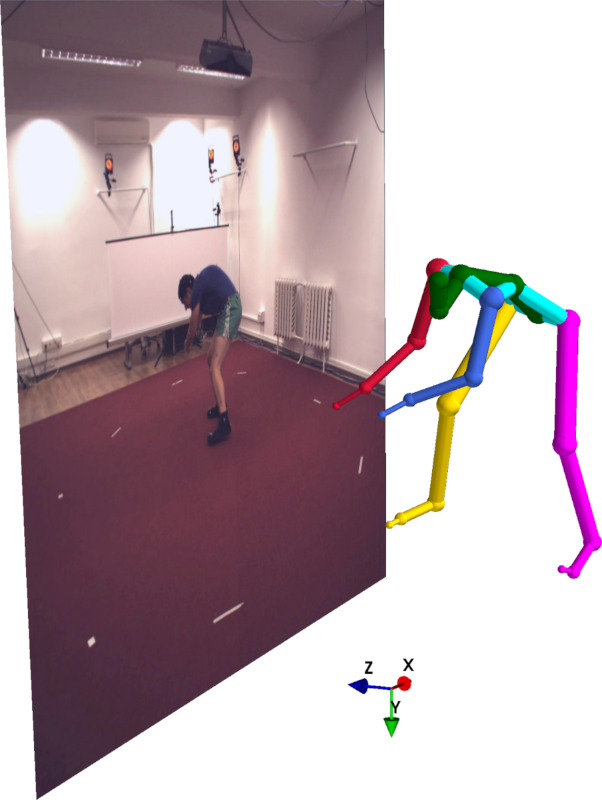}\hspace{0.05cm} 
  \includegraphics[height=2.23cm]{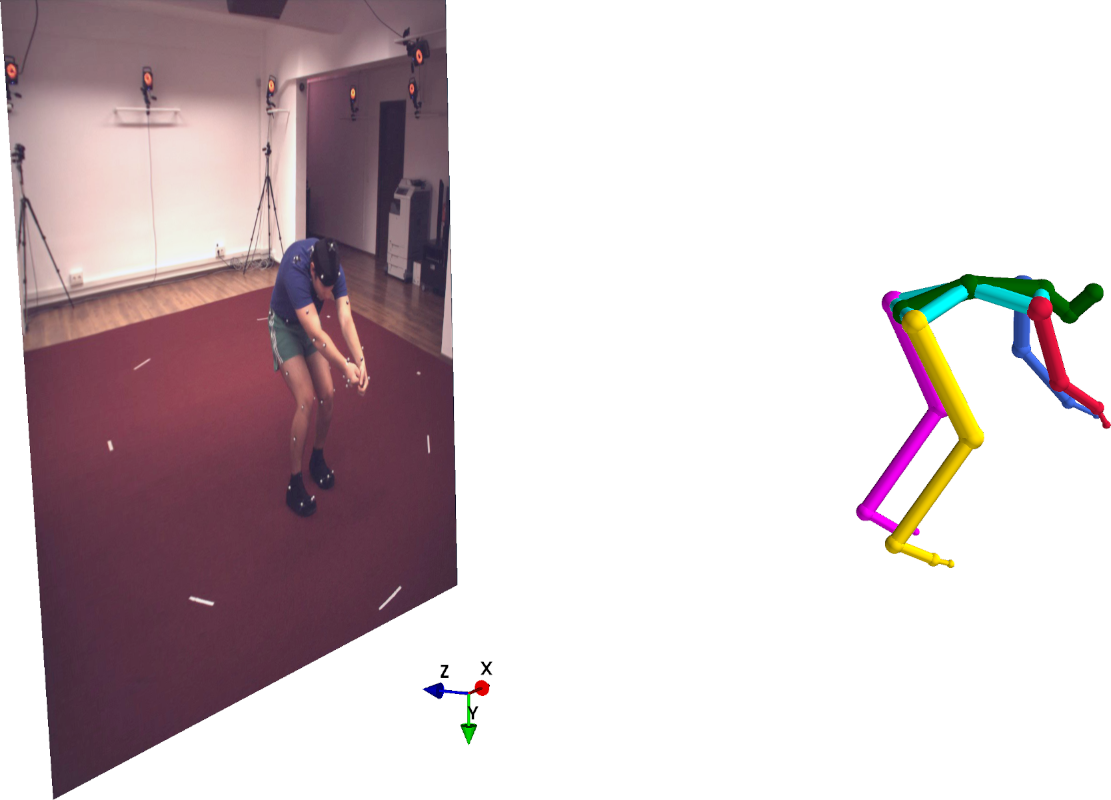}
  \caption{Absolute 3D pose predictions from monocular single images by our method.
  }
  \label{fig:predictions}
\end{figure}

\subsection{Qualitative results}

\begin{figure*}[!h]
  \centering
  \includegraphics[height=2.1cm]{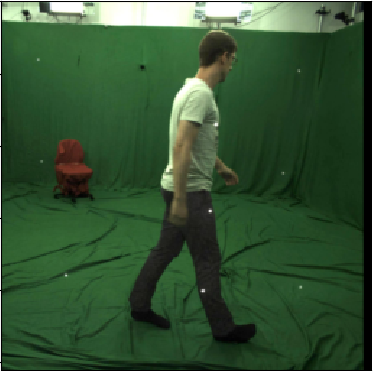}
  \includegraphics[height=2.2cm]{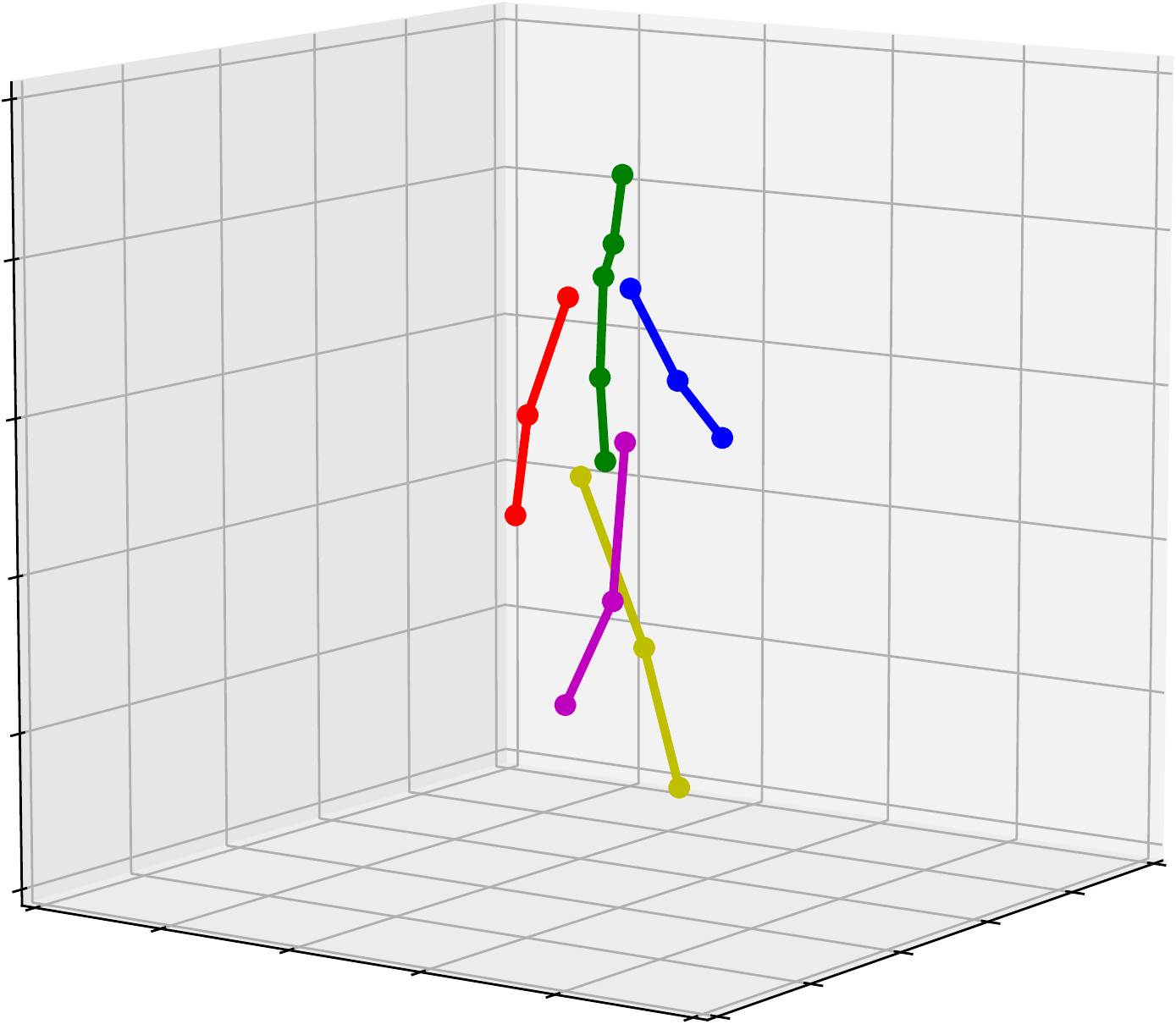}\hspace{0.3cm}
  \includegraphics[height=2.1cm]{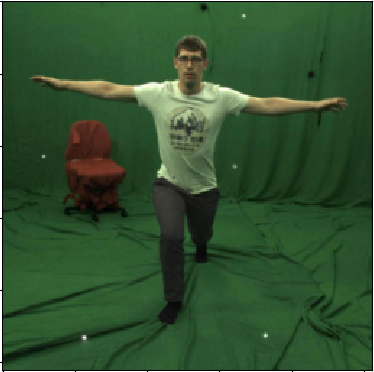}
  \includegraphics[height=2.2cm]{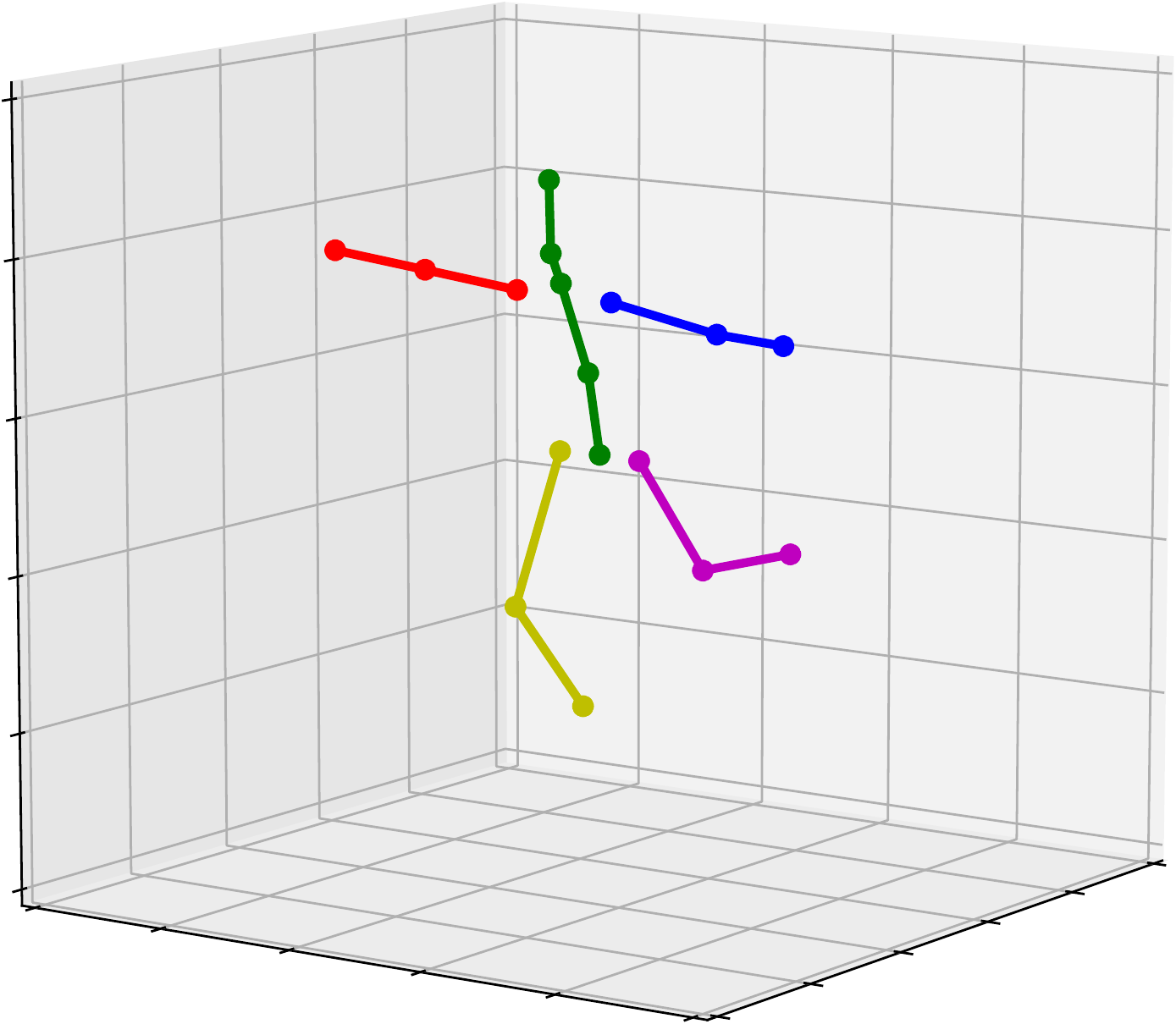}\hspace{0.3cm}
  \includegraphics[height=2.1cm]{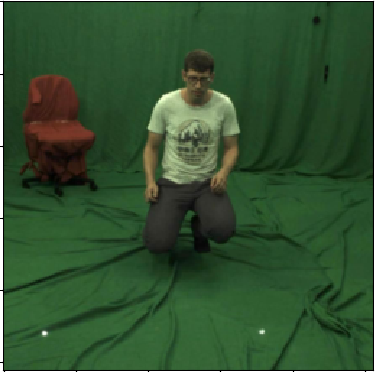}
  \includegraphics[height=2.2cm]{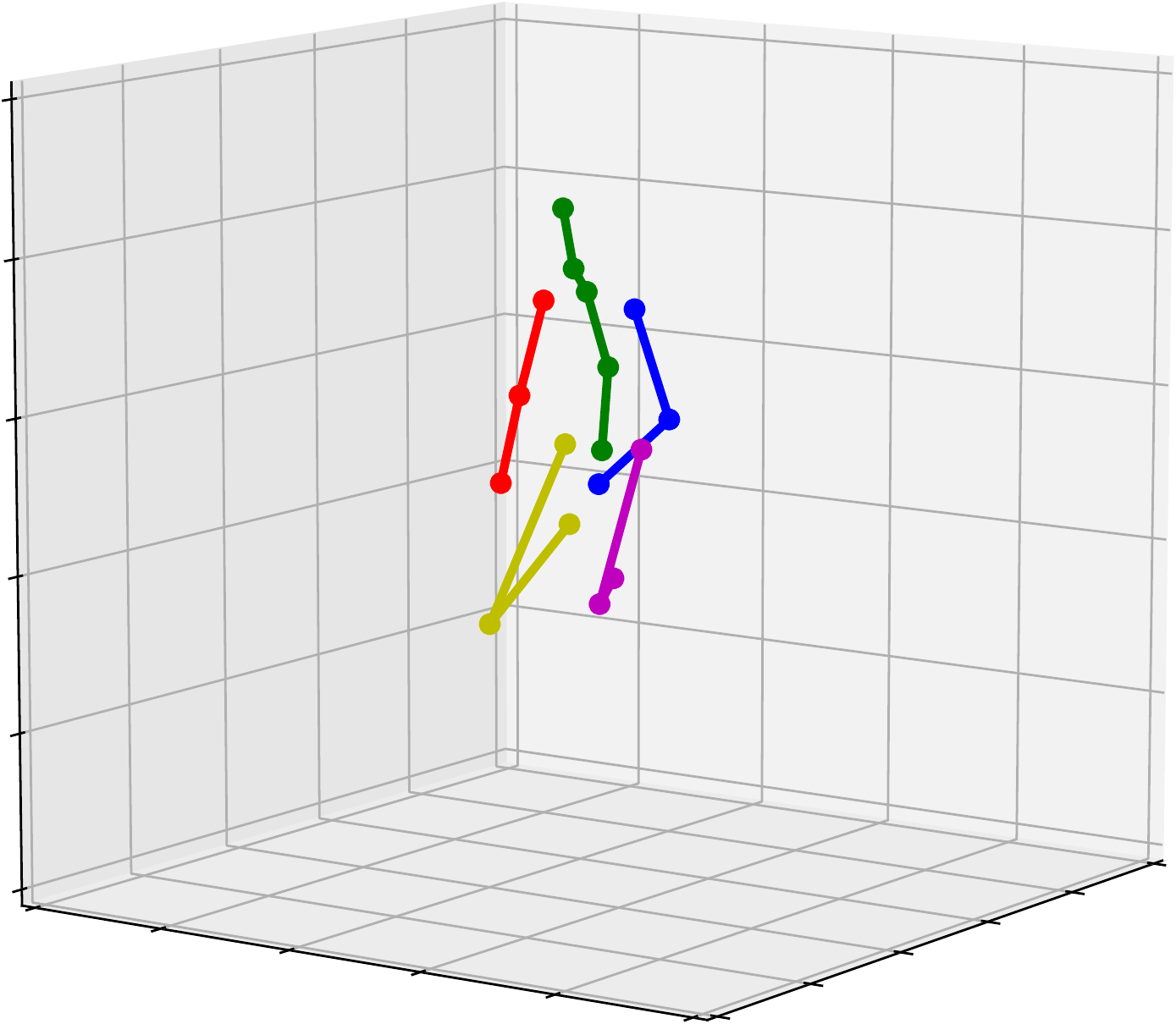}\\\vspace{0.01cm}
  \includegraphics[height=2.1cm]{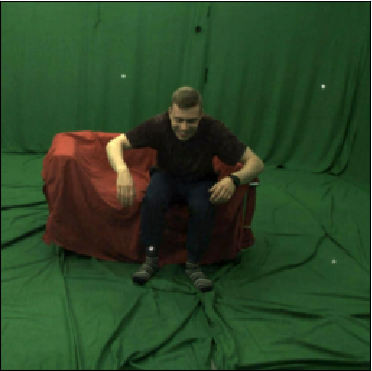}
  \includegraphics[height=2.2cm]{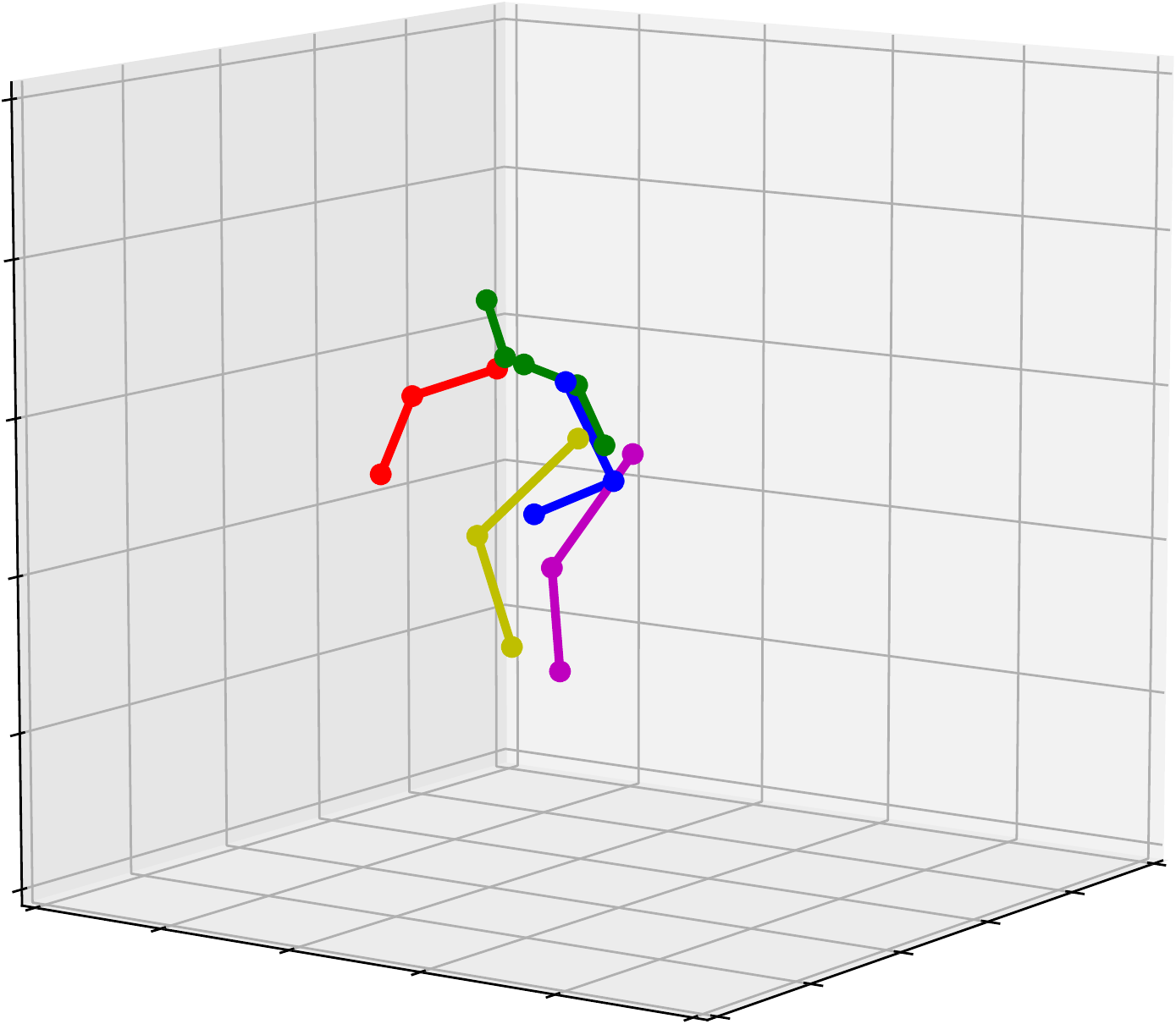}\hspace{0.3cm}
  \includegraphics[height=2.1cm]{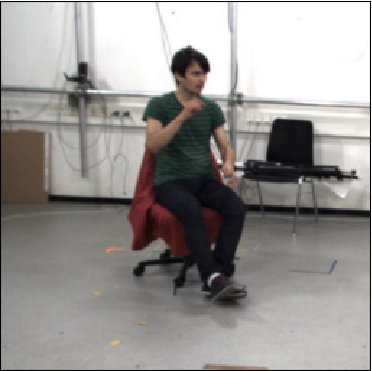}
  \includegraphics[height=2.2cm]{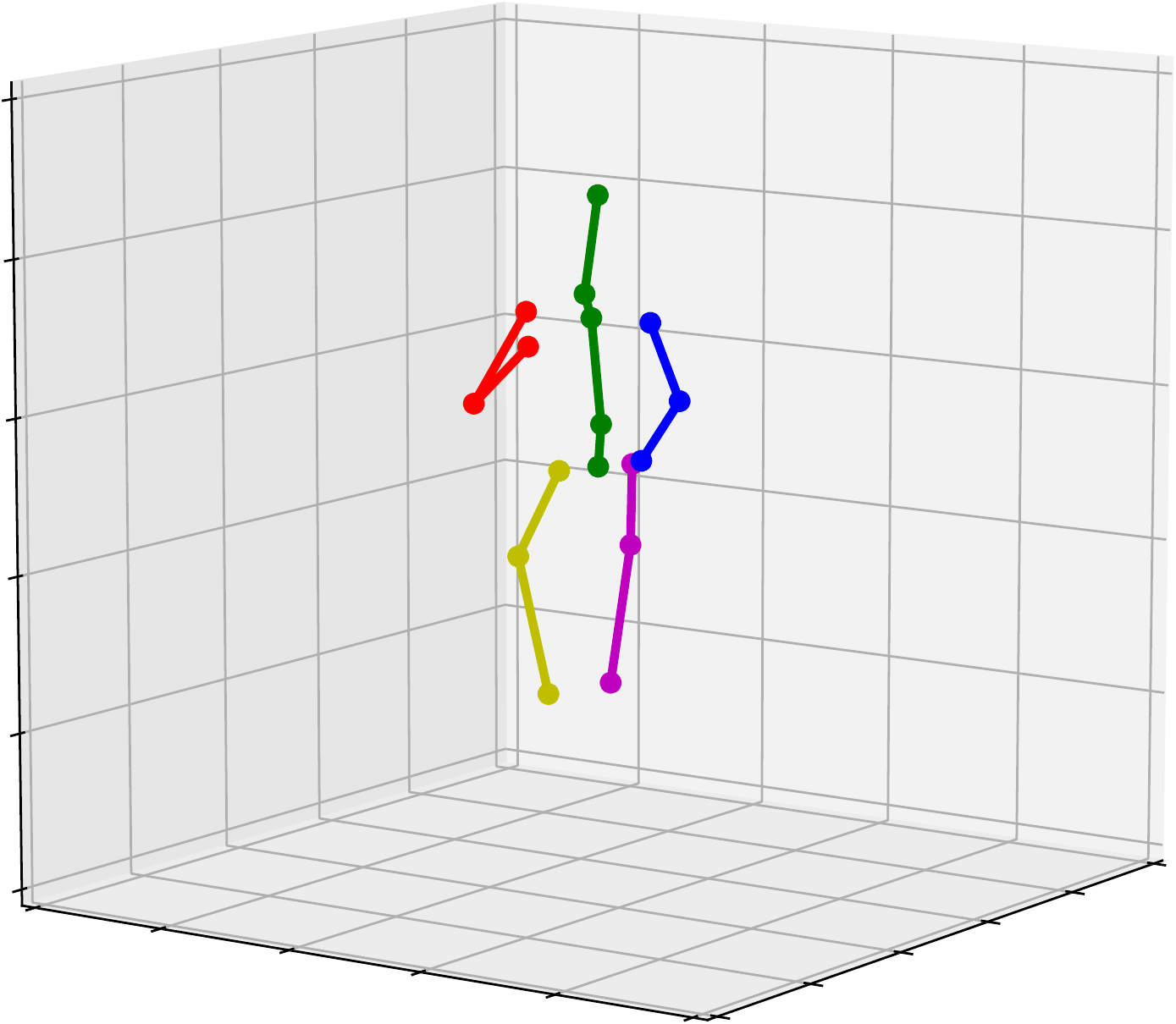}\hspace{0.3cm}
  \includegraphics[height=2.1cm]{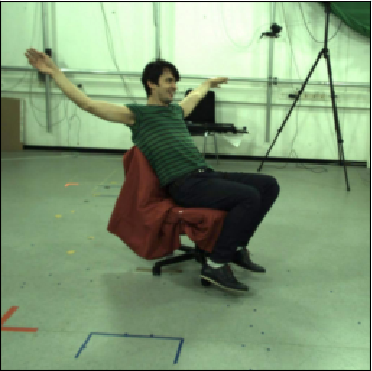}
  \includegraphics[height=2.2cm]{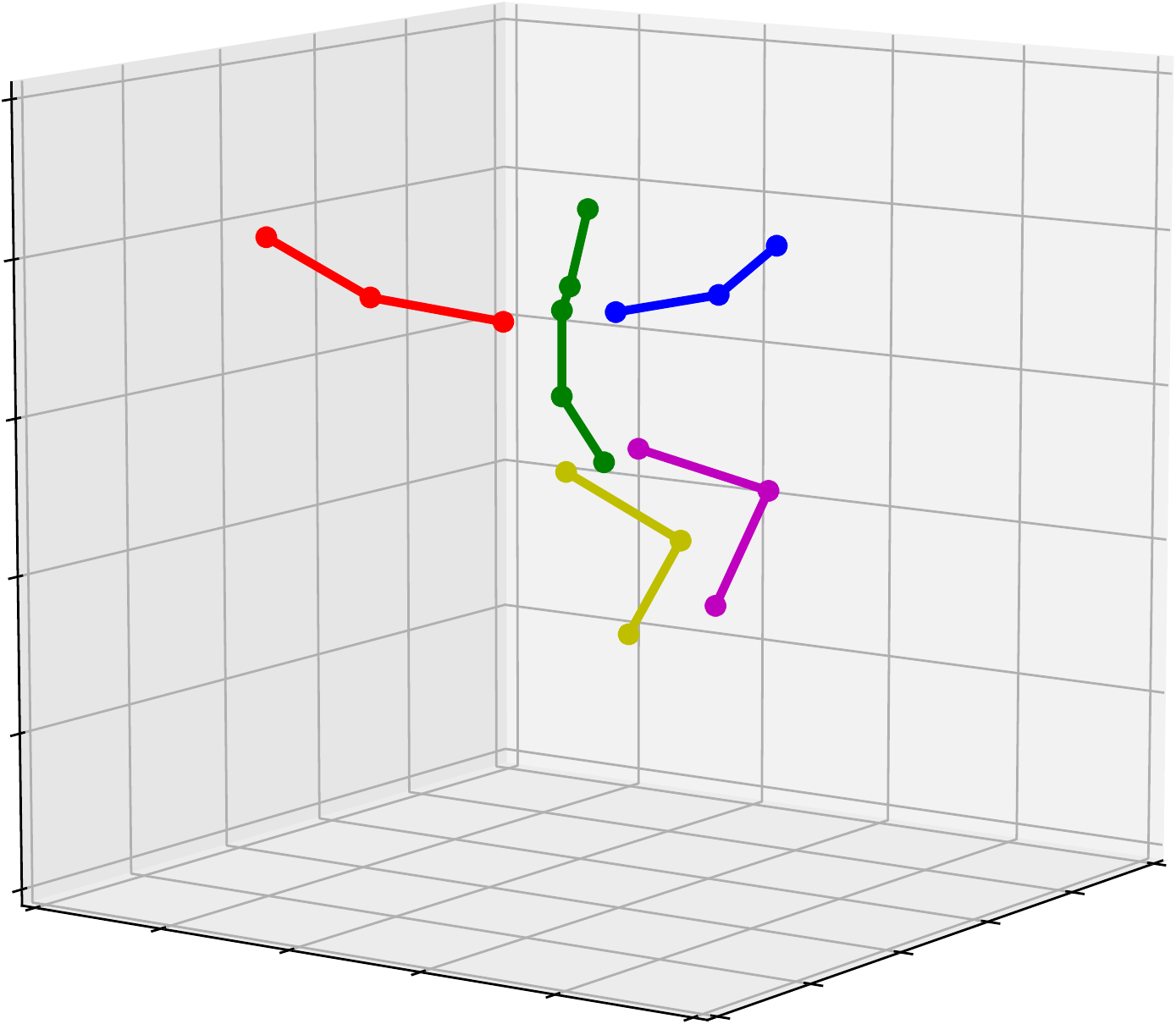}\\\vspace{0.01cm}
  \includegraphics[height=2.1cm]{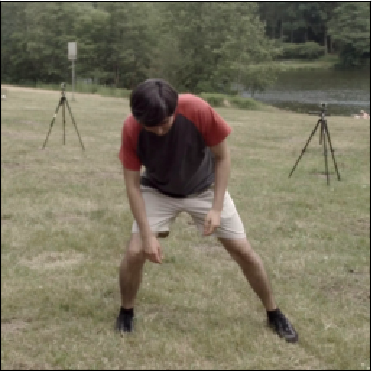}
  \includegraphics[height=2.2cm]{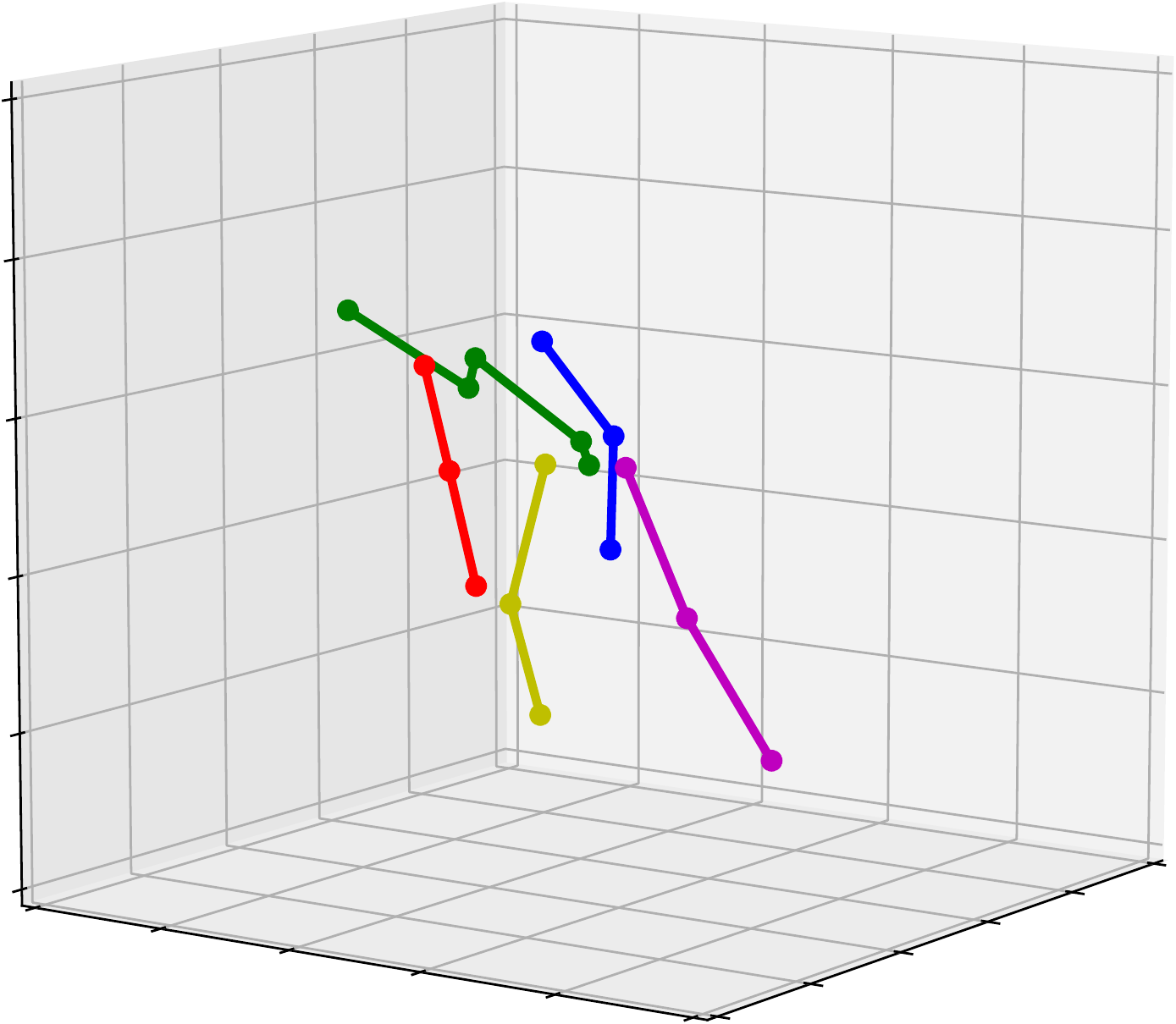}\hspace{0.3cm}
  \includegraphics[height=2.1cm]{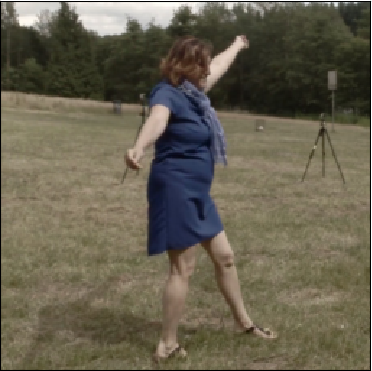}
  \includegraphics[height=2.2cm]{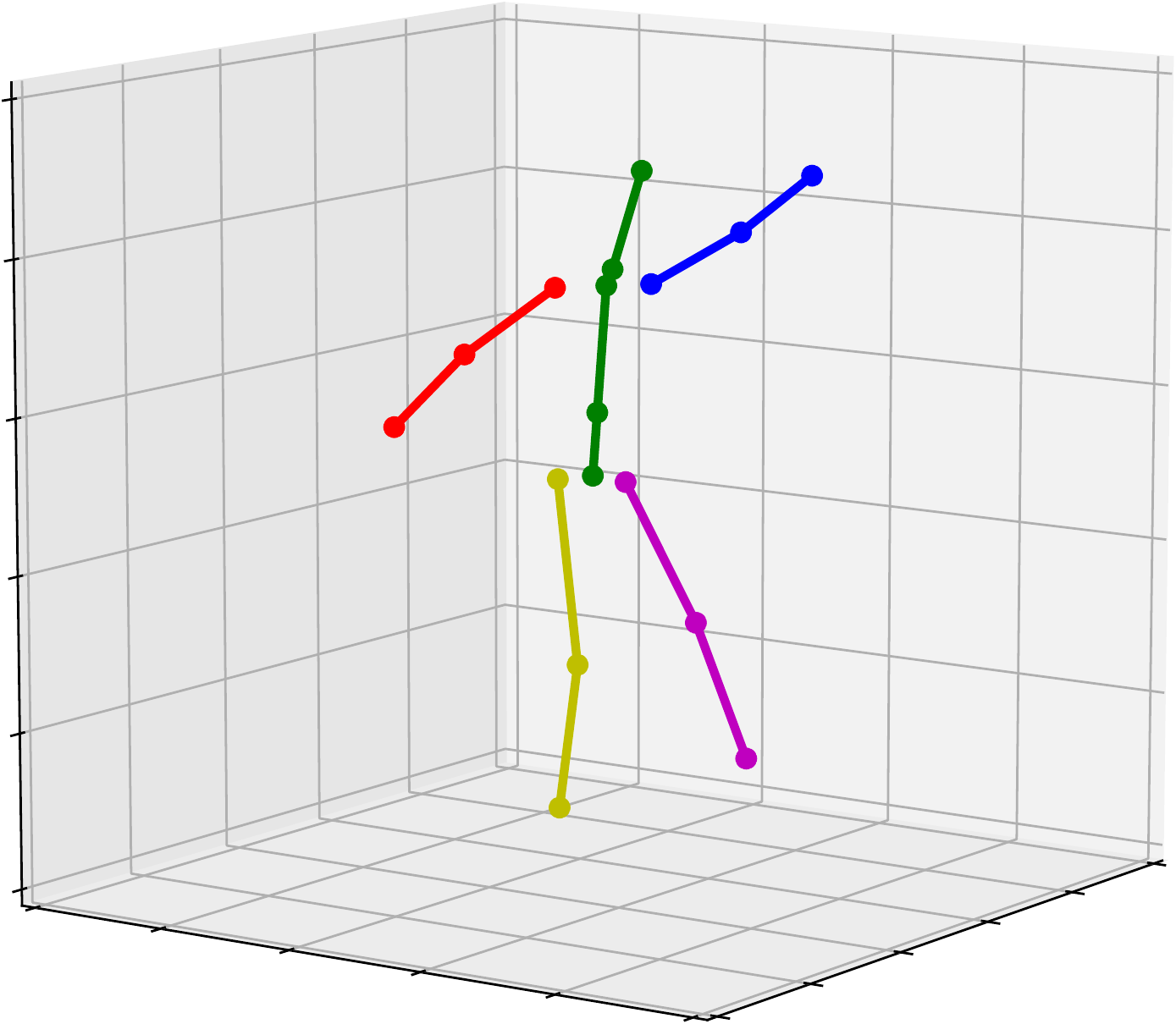}\hspace{0.3cm}
  \includegraphics[height=2.1cm]{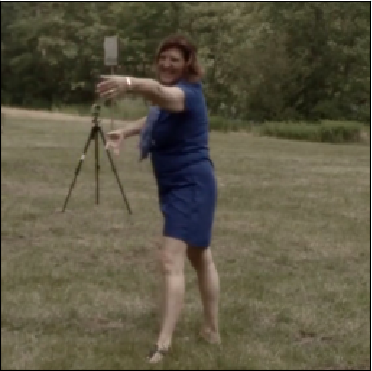}
  \includegraphics[height=2.2cm]{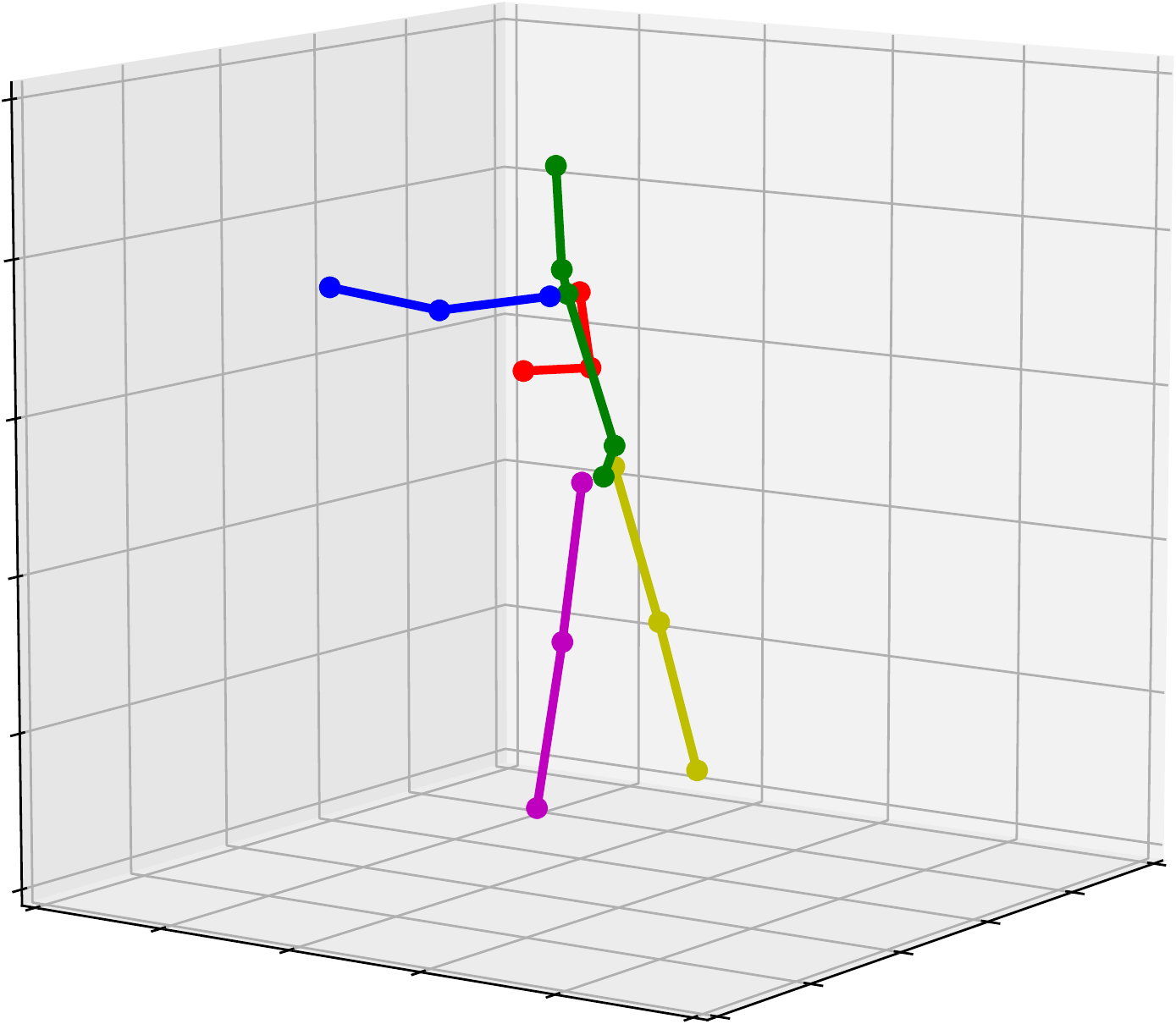}\\\vspace{0.01cm}
  \caption{
    3D pose predictions from monocular single images on MPI-INF-3DHP dataset,
    including indoor and outdoor scenes.
  }
  \label{fig:predictions_mpi}
\end{figure*}

\begin{figure*}[!h]
  \centering
  \includegraphics[height=2.1cm]{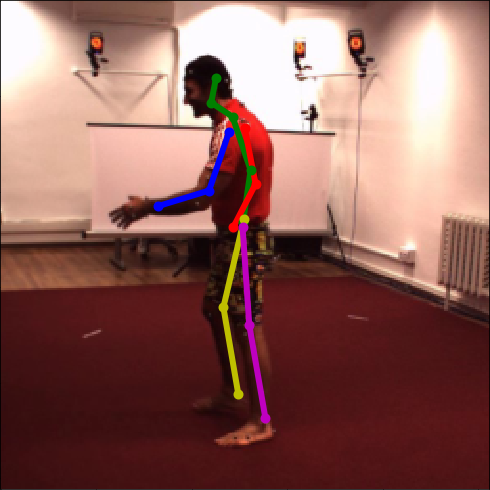}
  \includegraphics[height=2.1cm]{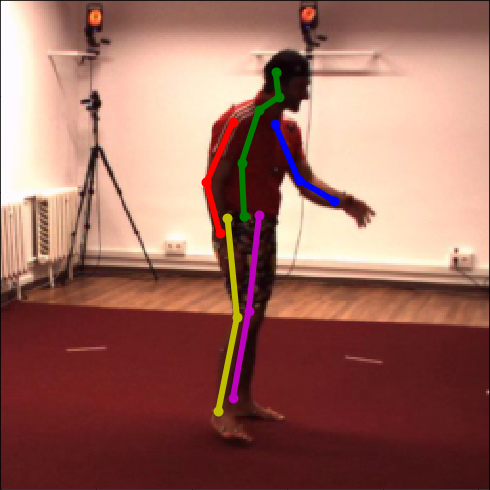}
  \includegraphics[height=2.1cm]{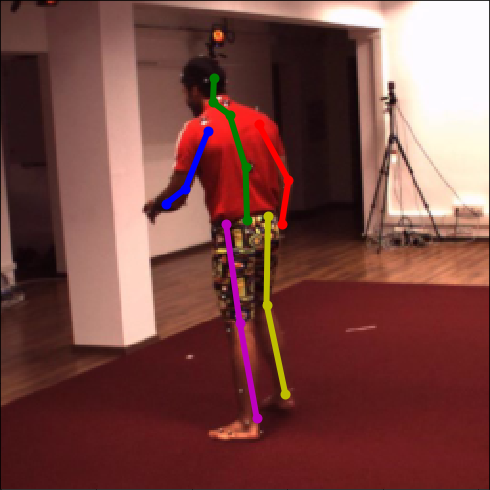}
  \includegraphics[height=2.1cm]{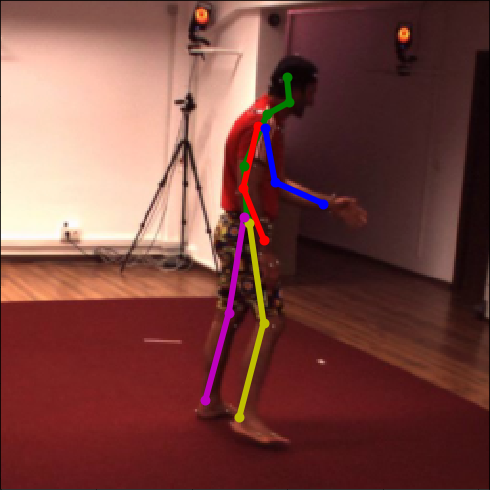}\hspace{0.1cm}
  \includegraphics[height=2.2cm]{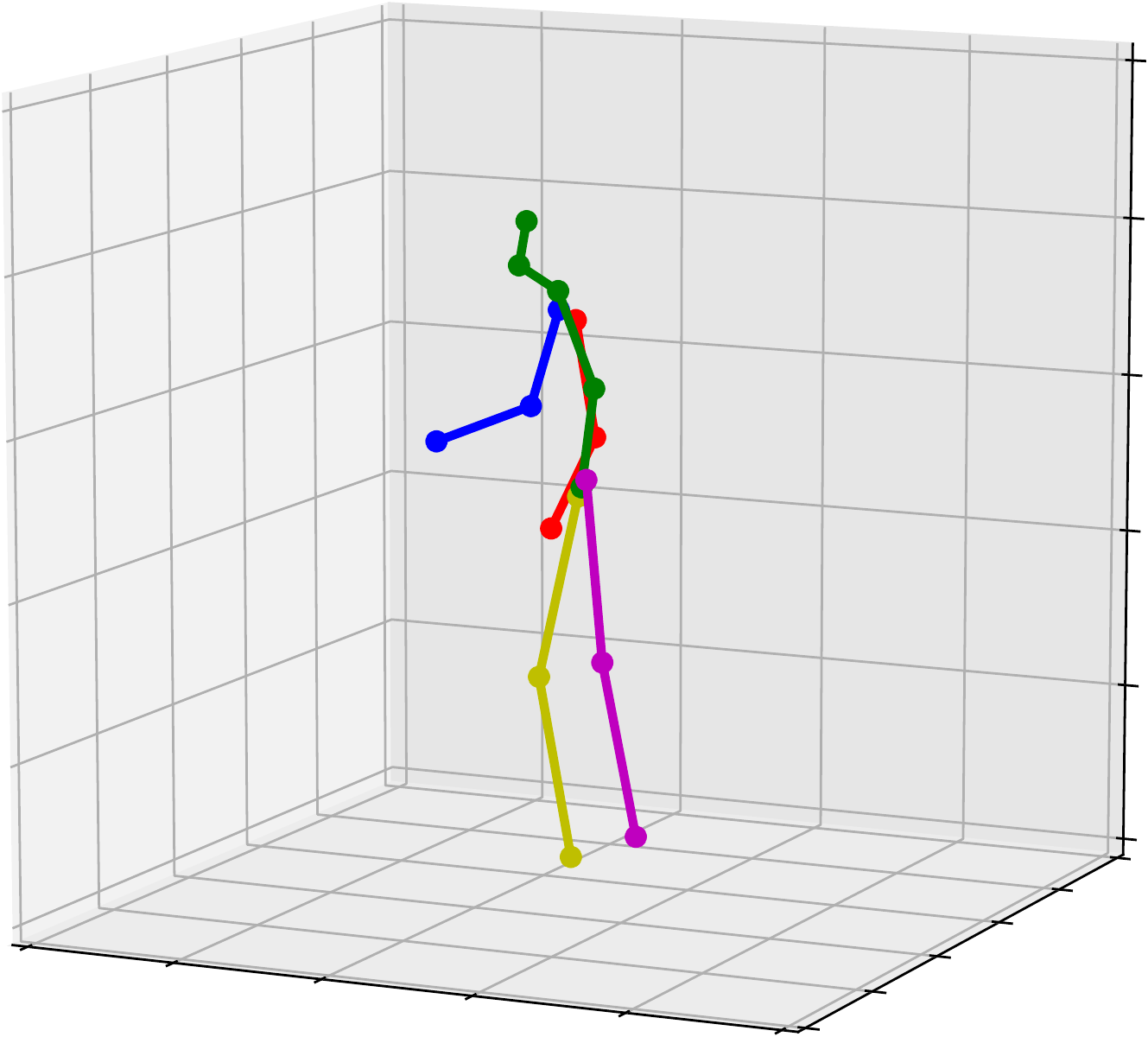}
  \includegraphics[height=2.2cm]{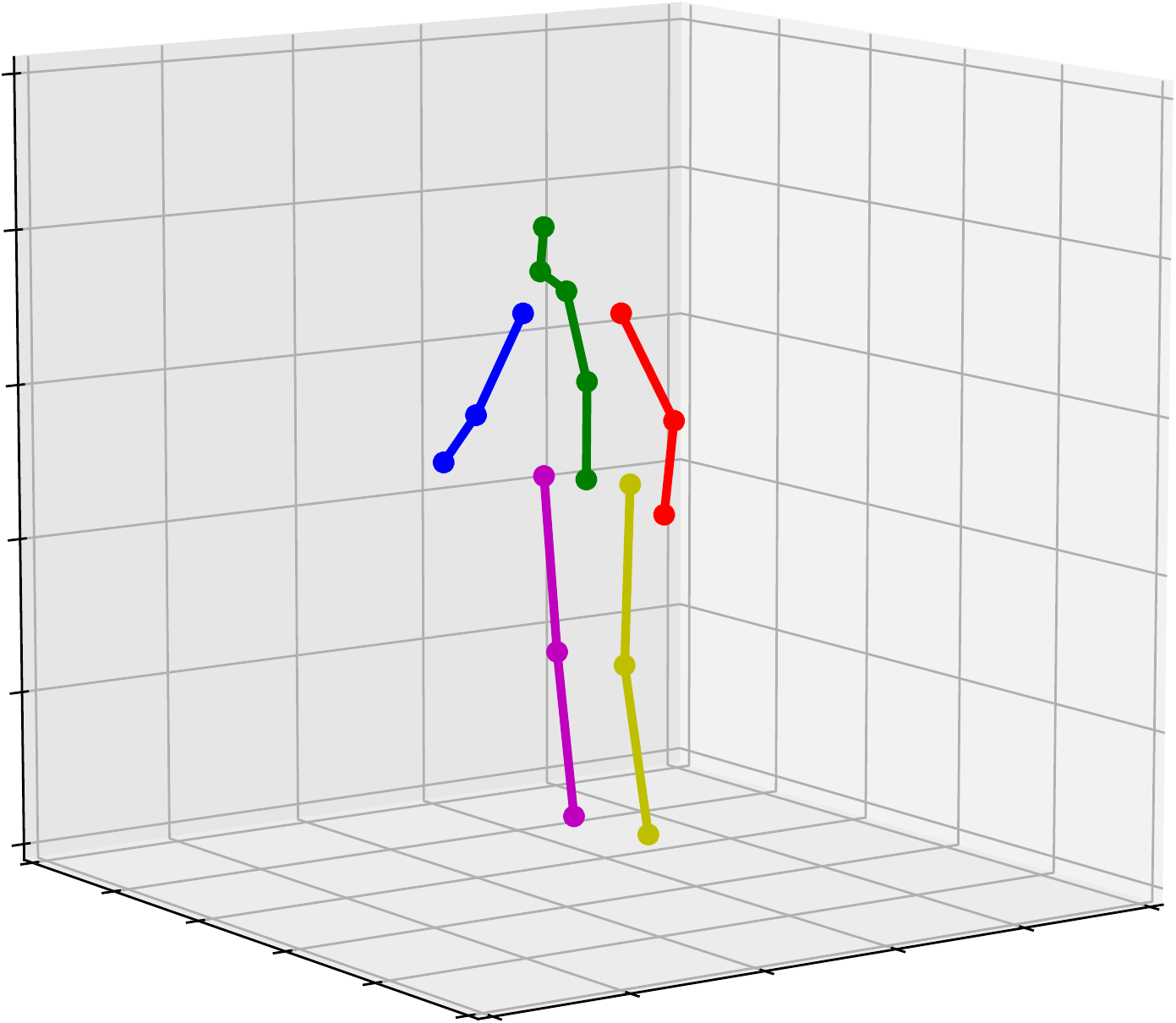}\\\vspace{0.01cm}
  \includegraphics[height=2.1cm]{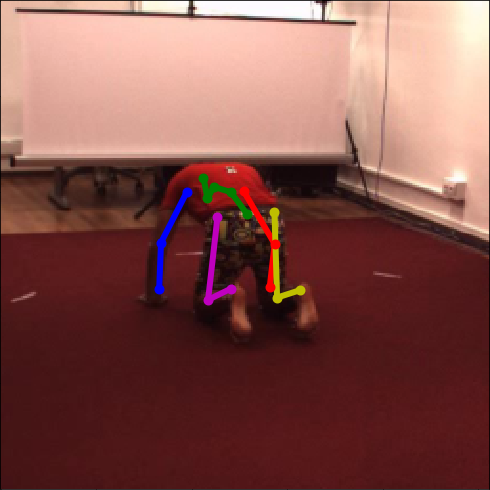}
  \includegraphics[height=2.1cm]{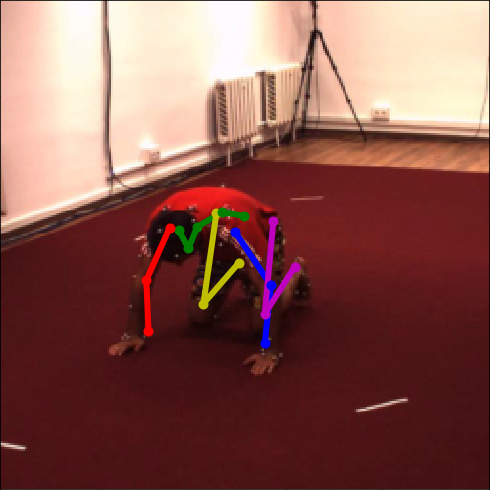}
  \includegraphics[height=2.1cm]{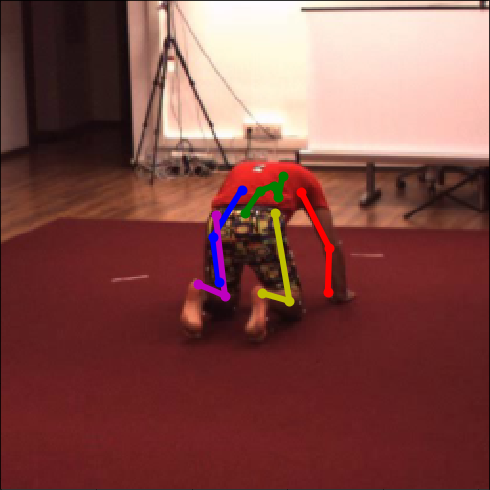}
  \includegraphics[height=2.1cm]{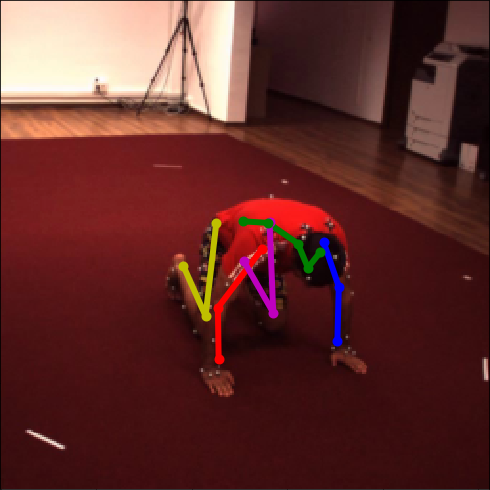}\hspace{0.1cm}
  \includegraphics[height=2.2cm]{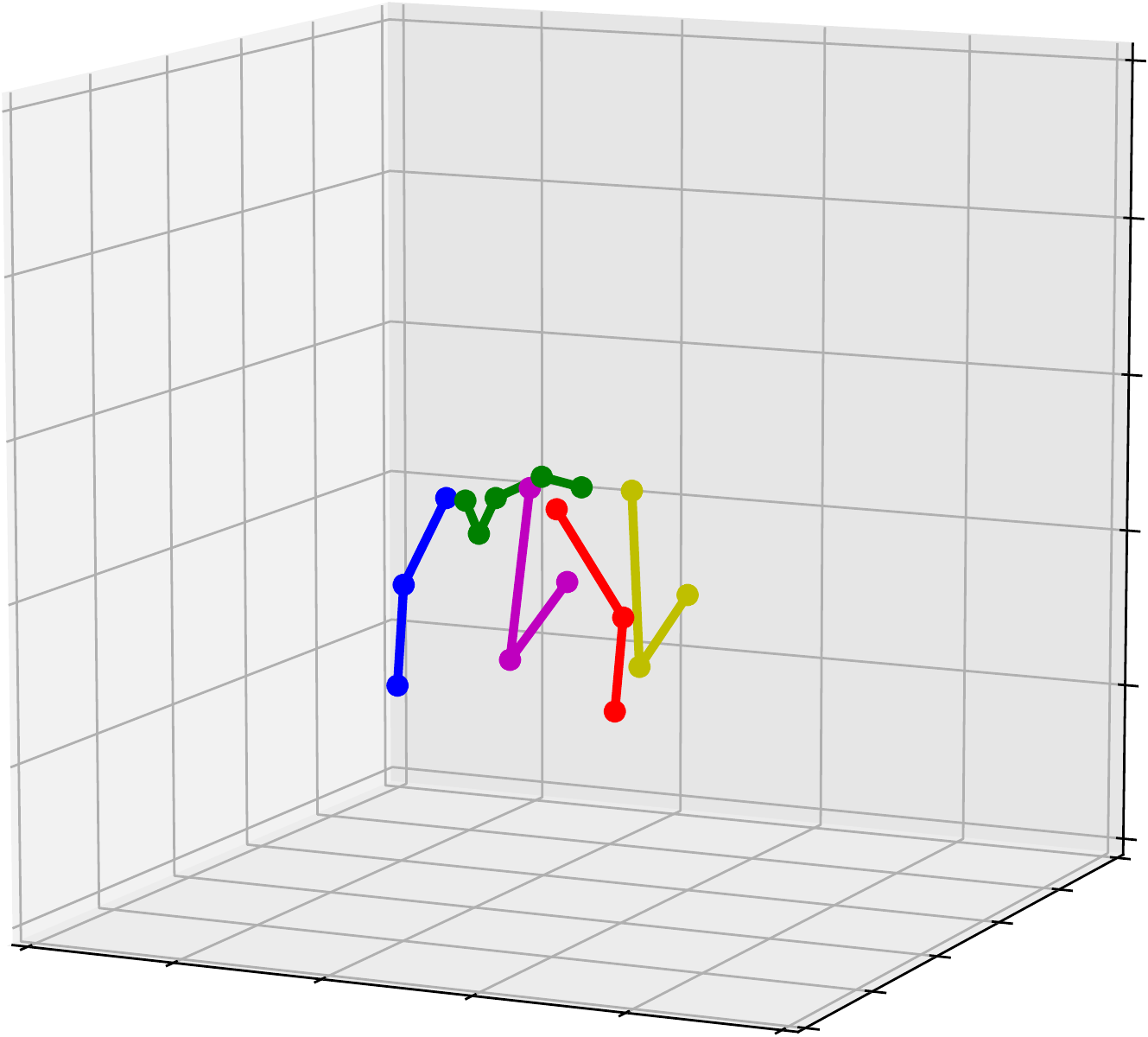}
  \includegraphics[height=2.2cm]{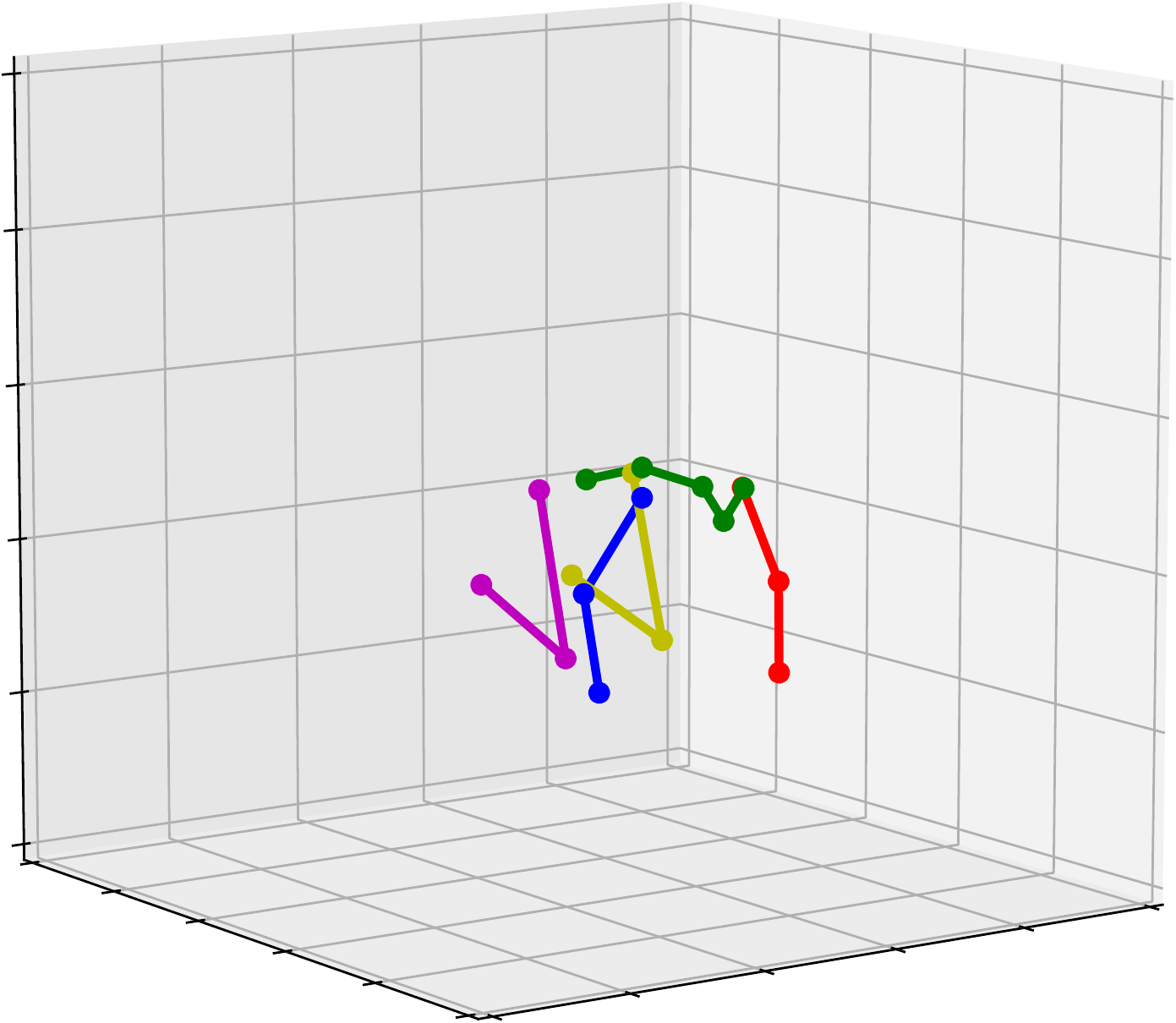}\\\vspace{0.01cm}
  \includegraphics[height=2.1cm]{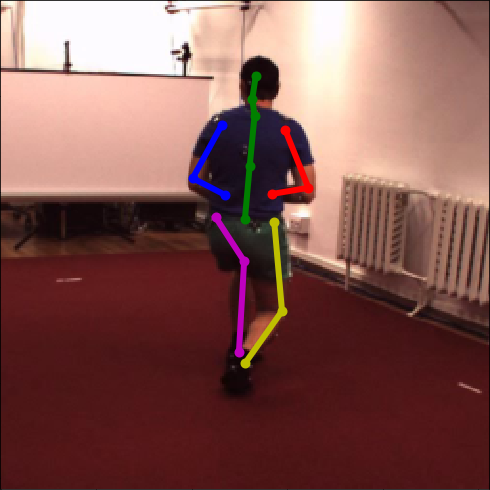}
  \includegraphics[height=2.1cm]{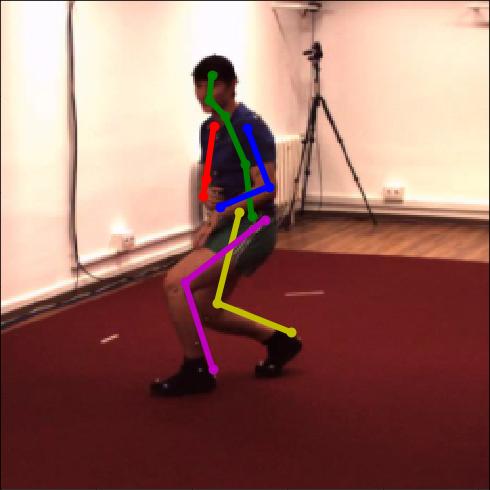}
  \includegraphics[height=2.1cm]{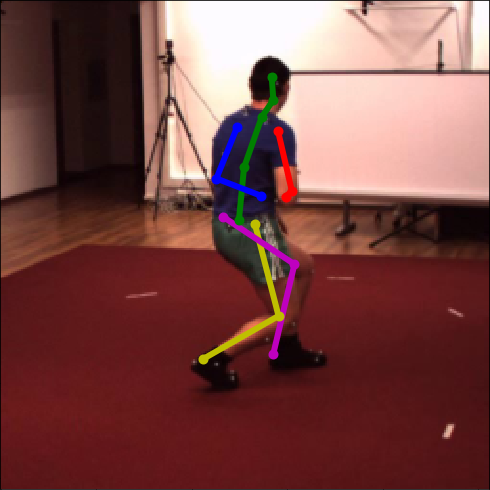}
  \includegraphics[height=2.1cm]{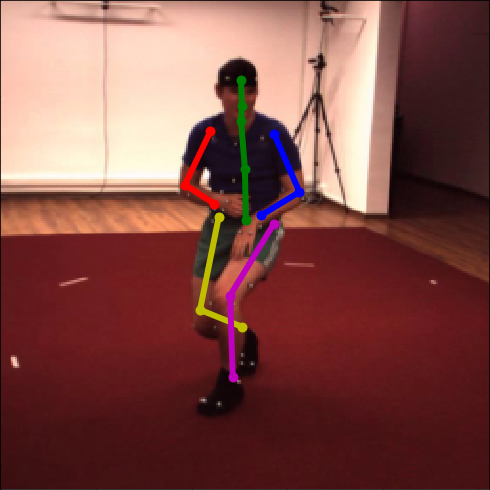}\hspace{0.1cm}
  \includegraphics[height=2.2cm]{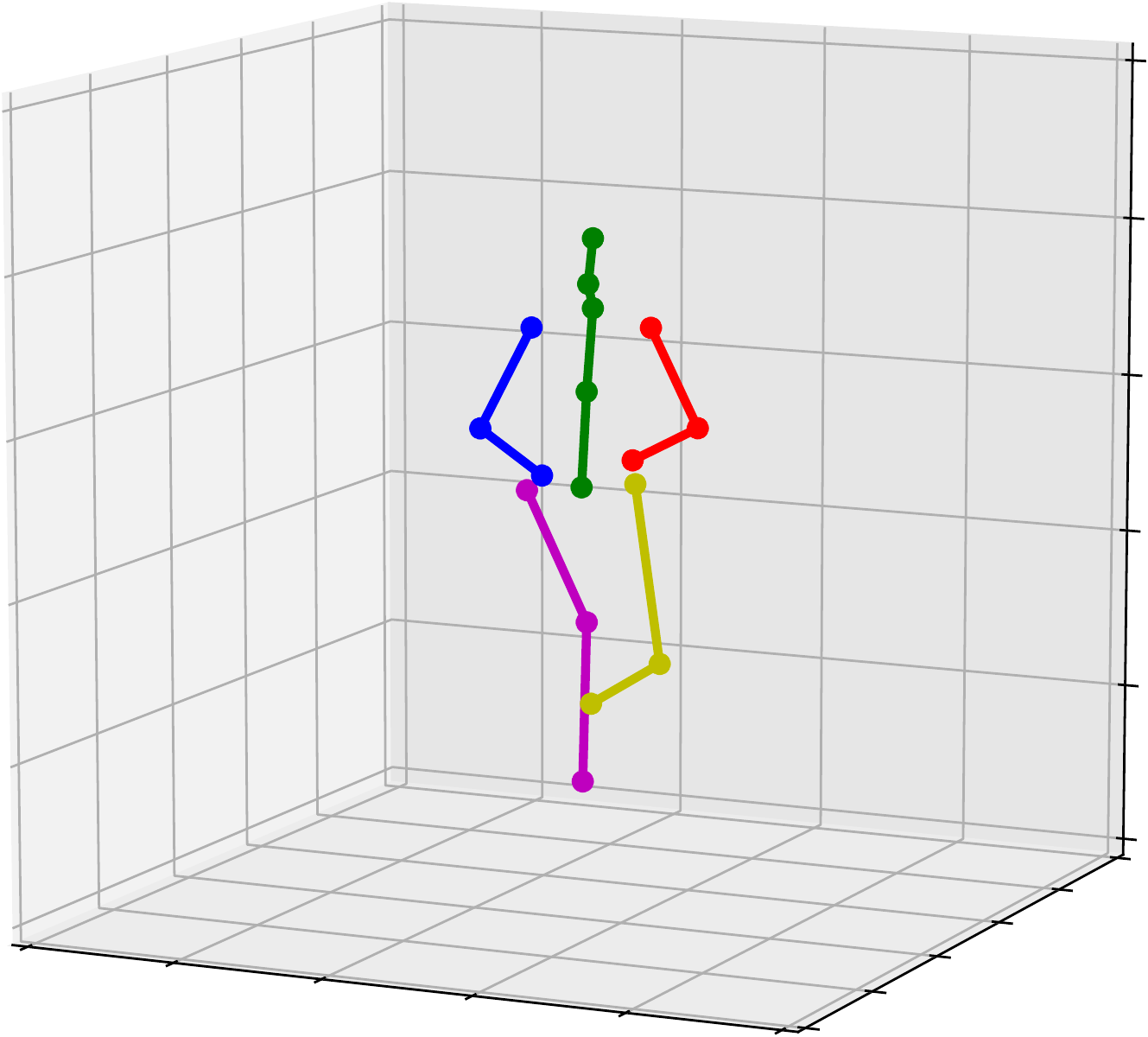}
  \includegraphics[height=2.2cm]{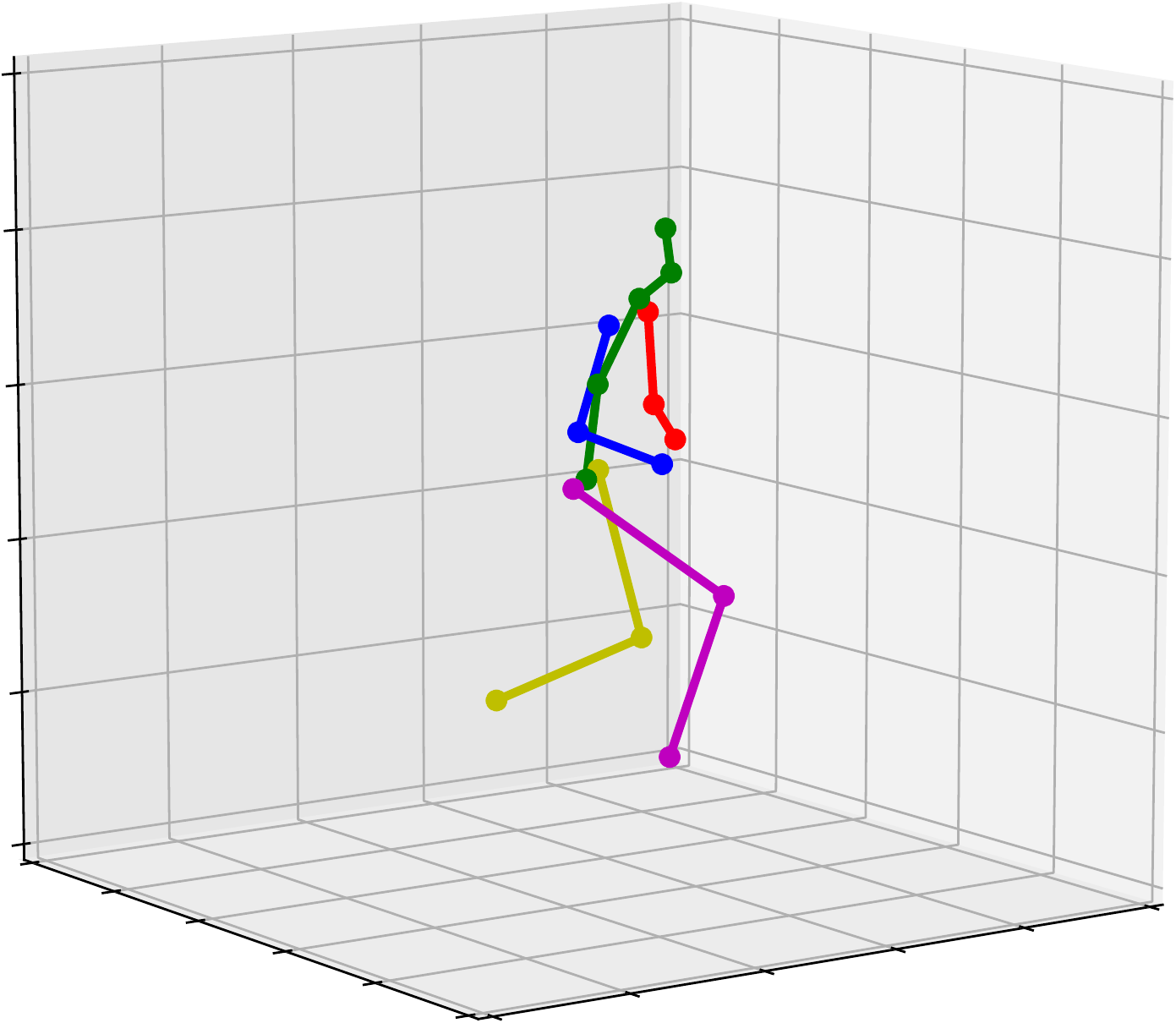}\\\vspace{0.01cm}
  \caption{
    3D pose predictions from our consensus-based optimization algorithm, considering multi-view on Human3.6M.
    Final 3D poses are projected into the different views (a,b,c,d) and shown in perspective (e,f).
  }
  \label{fig:predictions_h36m}
\end{figure*}

\begin{figure*}[!h]
  \centering
  \includegraphics[width=2.5cm]{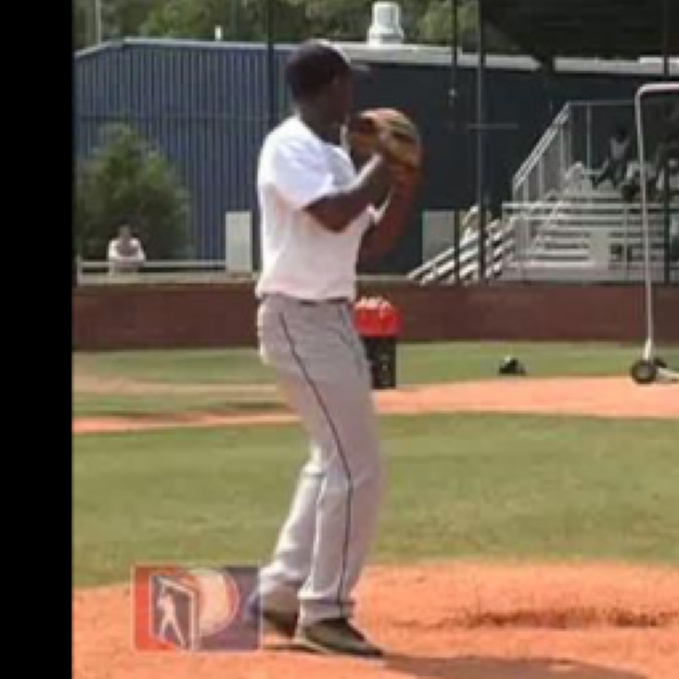}\hspace{0.2cm}
  \includegraphics[width=2.5cm]{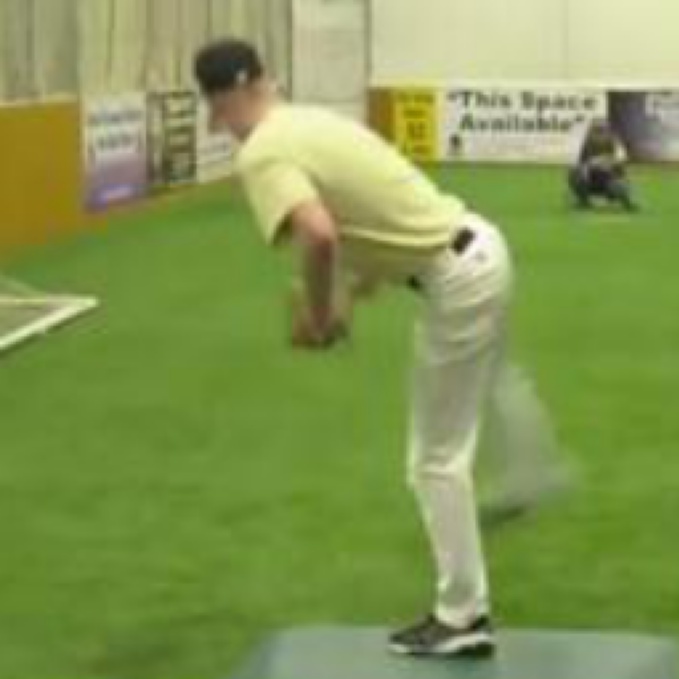}\hspace{0.2cm}
  \includegraphics[width=2.5cm]{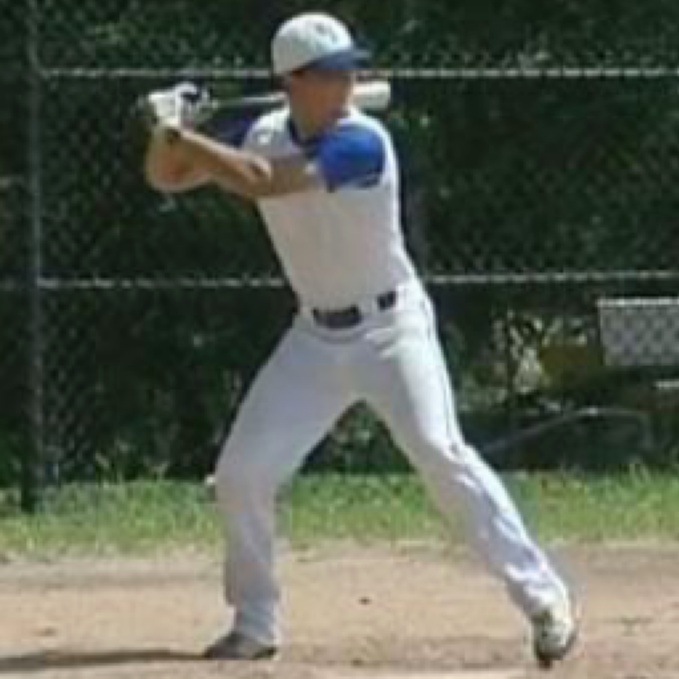}\hspace{0.2cm}
  \includegraphics[width=2.5cm]{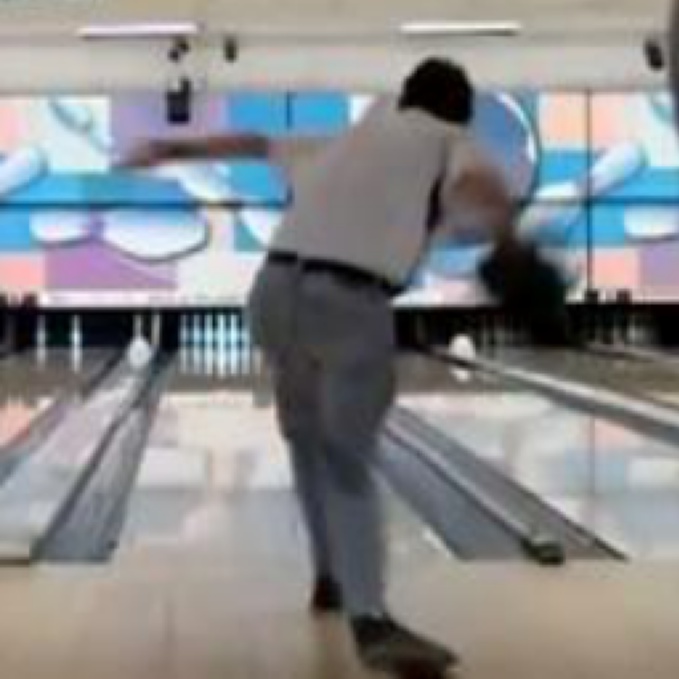}\hspace{0.2cm}
  \includegraphics[width=2.5cm]{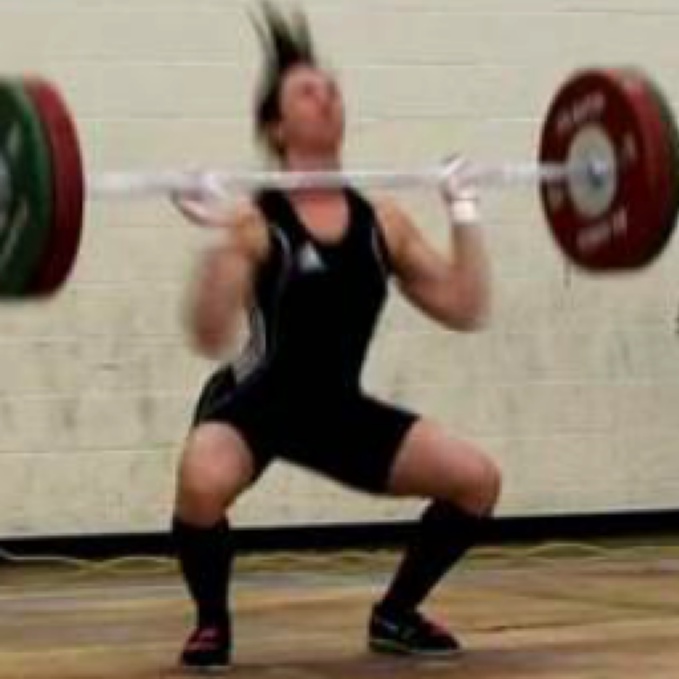}\hspace{0.2cm}
  \includegraphics[width=2.5cm]{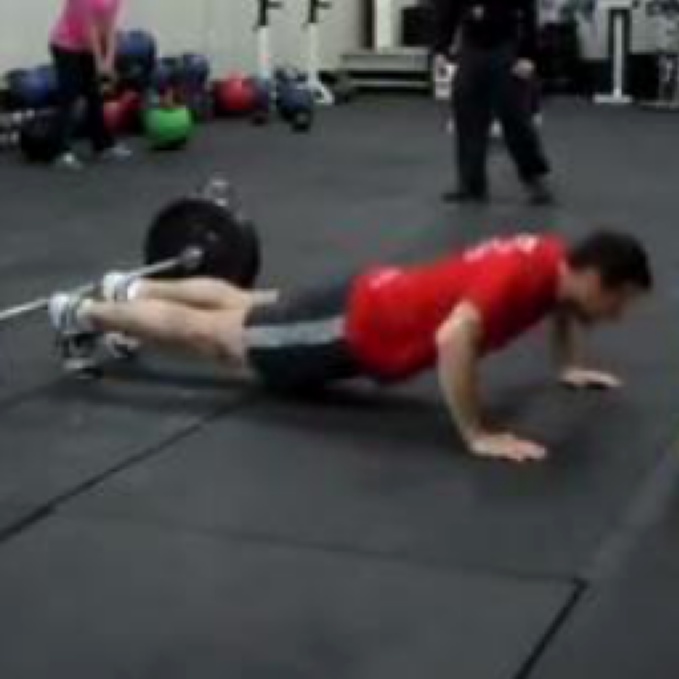}\\\vspace{0.2cm}
  \includegraphics[width=2.7cm]{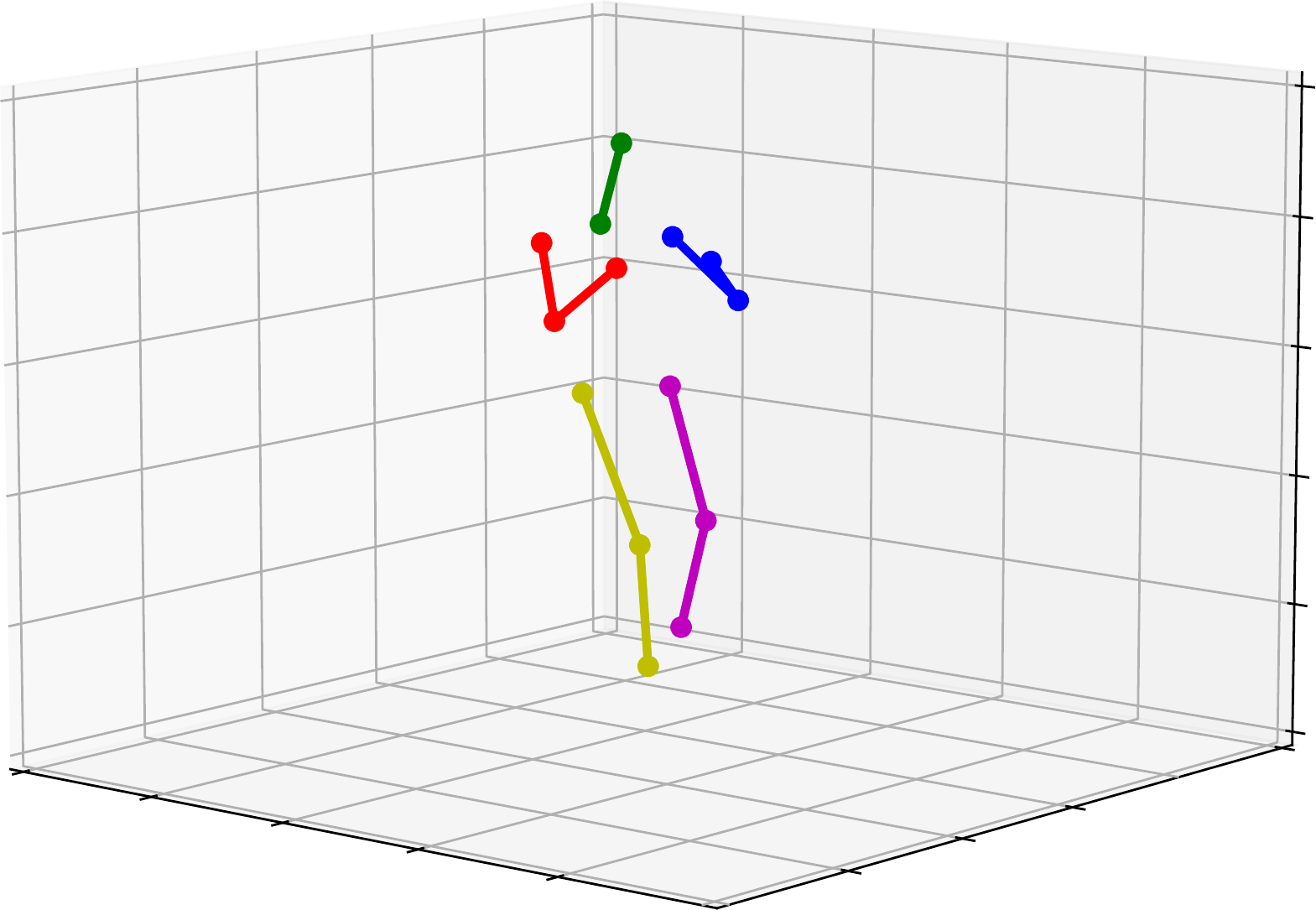}
  \includegraphics[width=2.7cm]{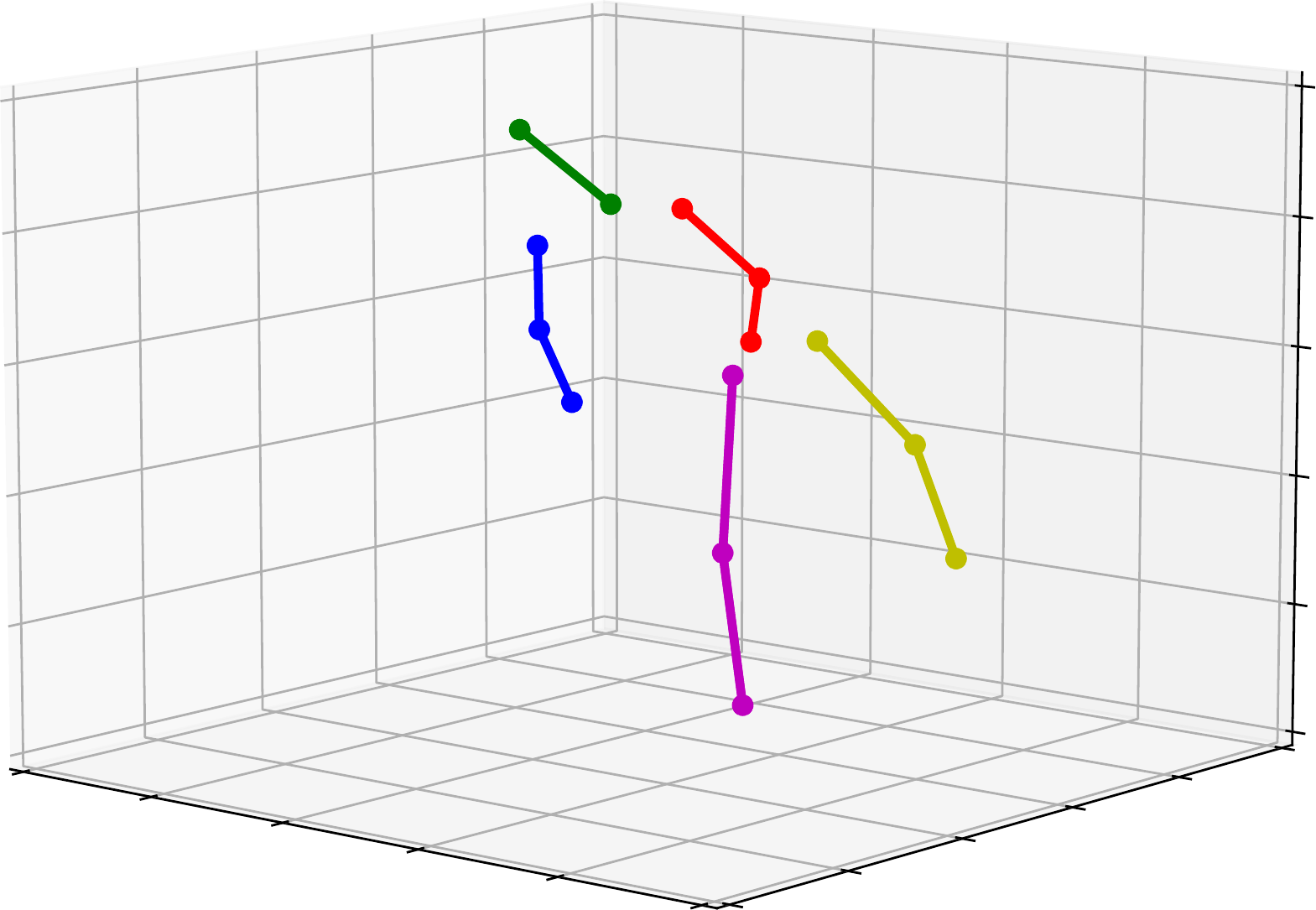}
  \includegraphics[width=2.7cm]{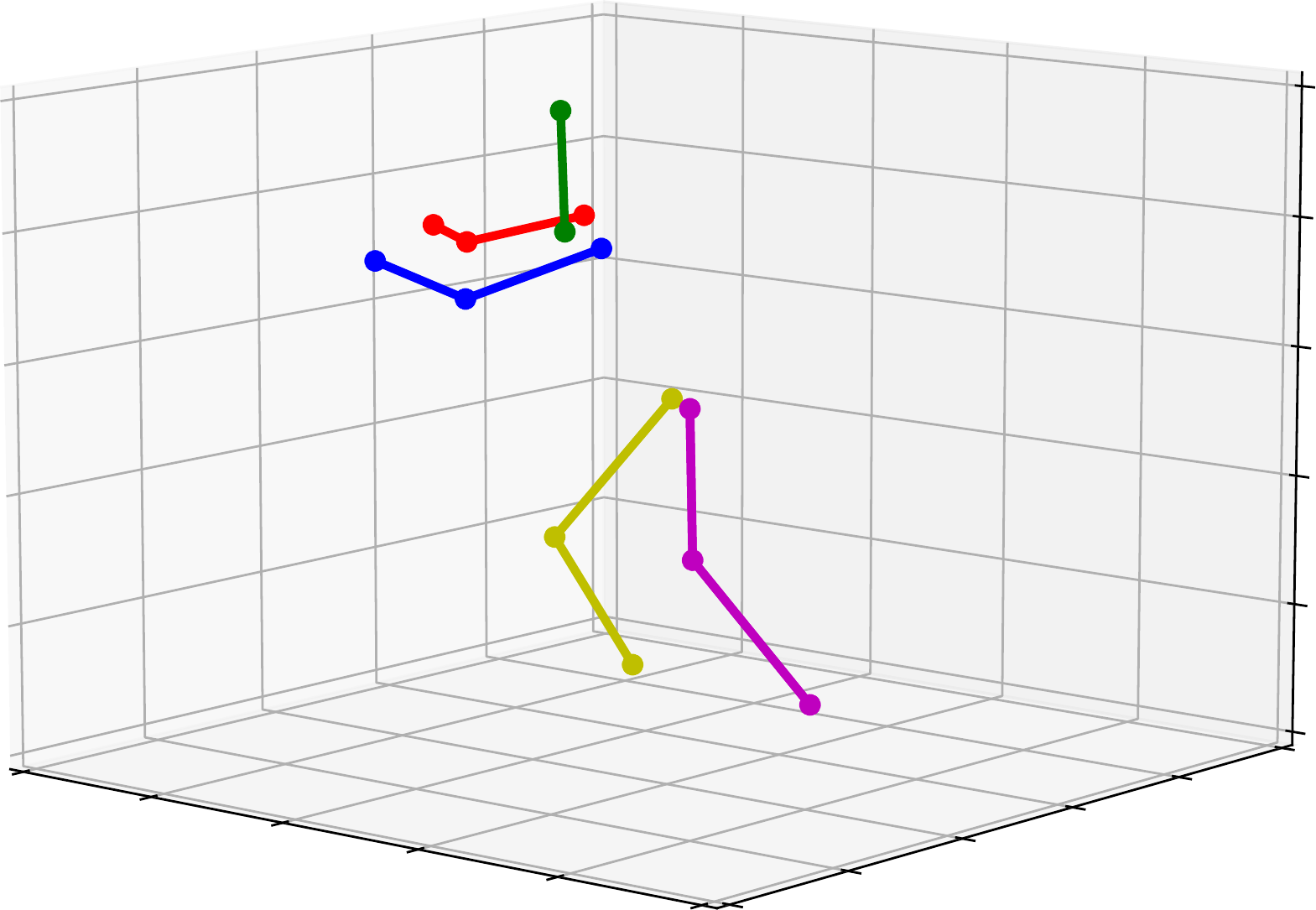}
  \includegraphics[width=2.7cm]{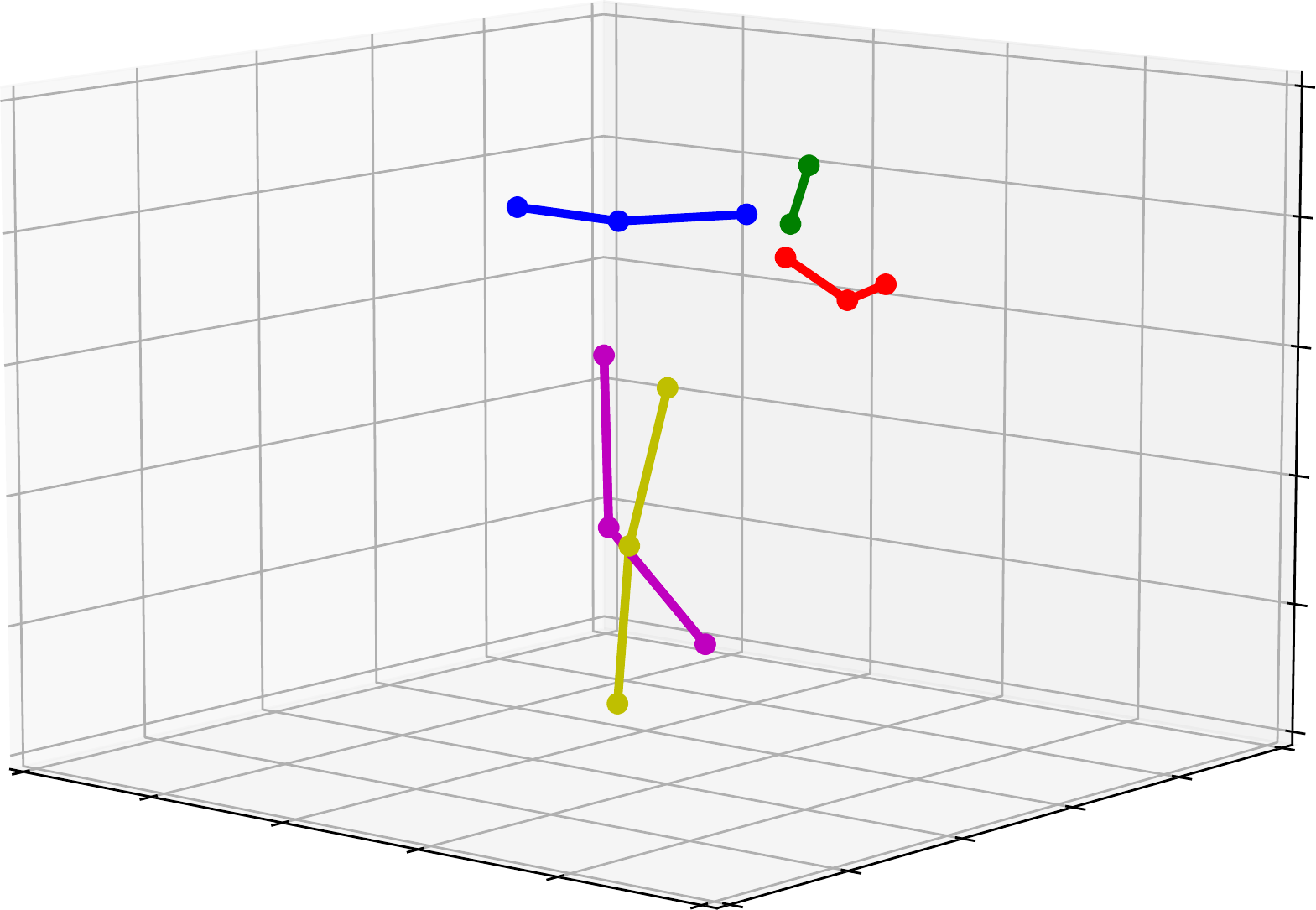}
  \includegraphics[width=2.7cm]{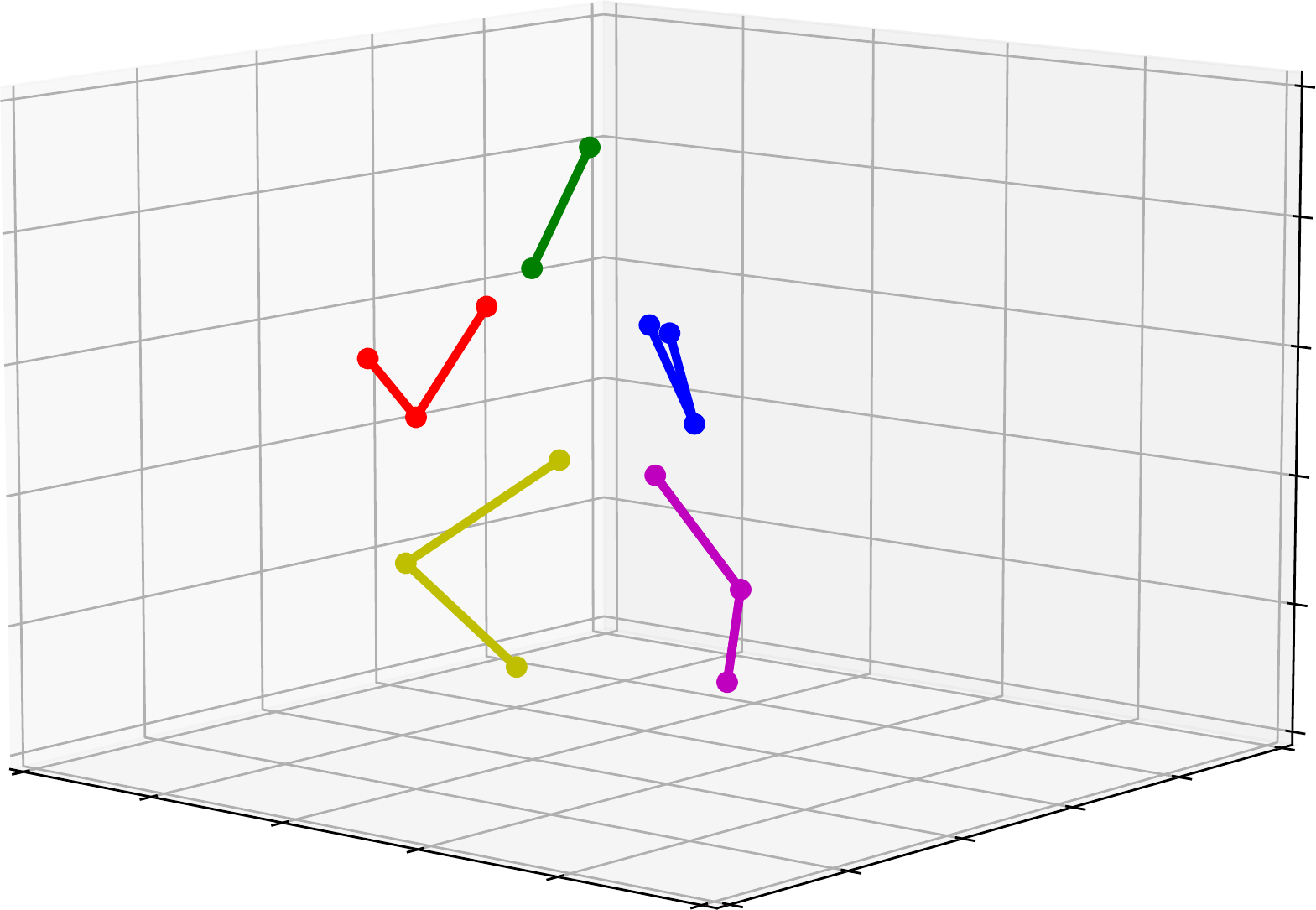}
  \includegraphics[width=2.7cm]{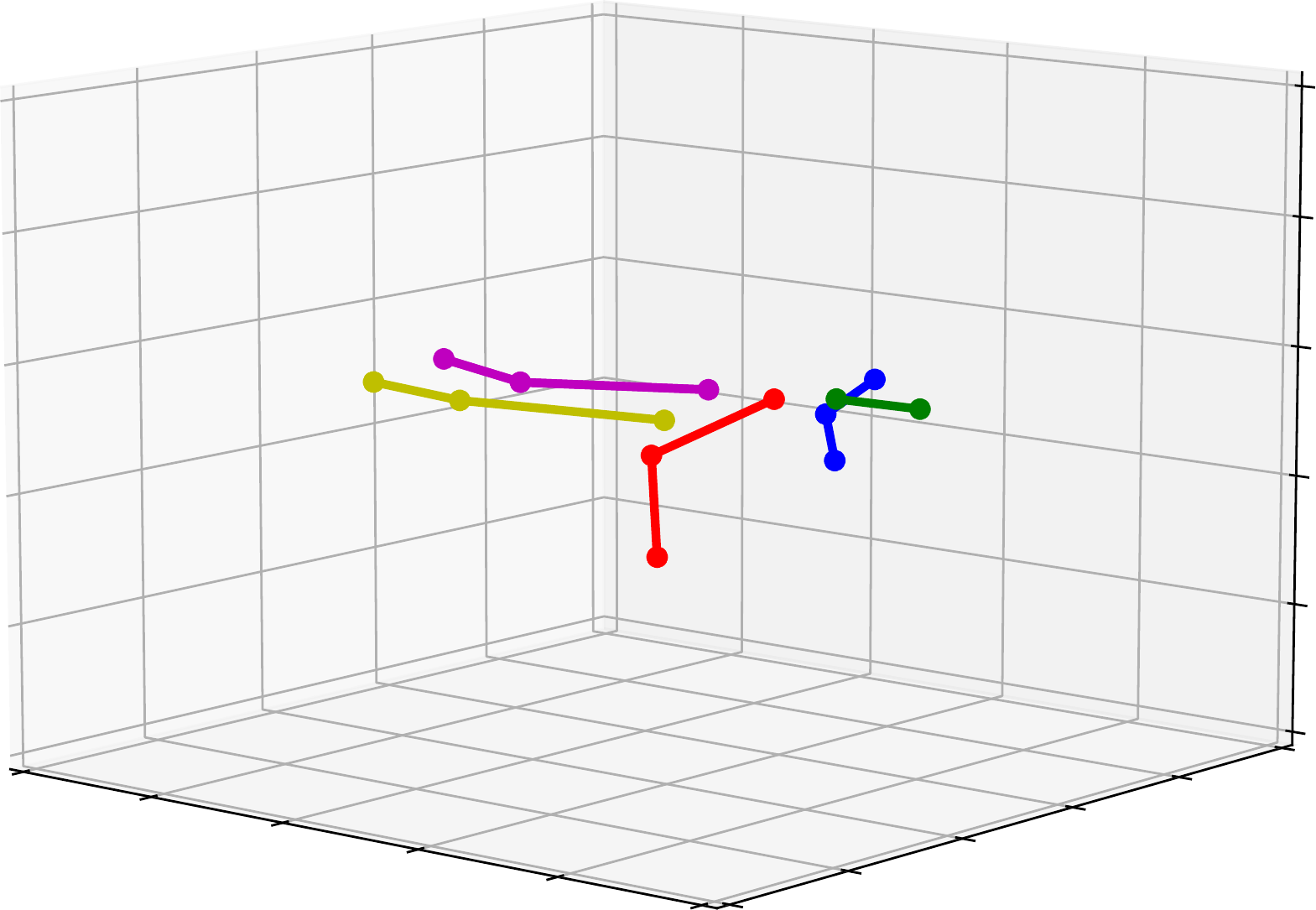}
  \caption{
    Generalization of our method for 3D pose estimation on unseen dataset (PennAction), including outdoor scenes in different contexts.
  }
  \label{fig:predictions_pennaction}
\end{figure*}

In Fig.~\ref{fig:predictions} we present some qualitative results of
predicted absolute 3D poses by our method. Not that the distance
from predictions to the images are proportional to the absolute distance in $z$.
In Fig.~\ref{fig:predictions_mpi} we show monocular predictions by our method
on the MPI-INF-3DHP dataset, including challenging outdoor scenes, which are
not present in the training set.
Finally, in Fig.~\ref{fig:predictions_h36m}, we show the
results from our consensus-based optimization approach, from multi-view
predictions on Human3.6M. \revb{Finally, in Fig.~\ref{fig:predictions_pennaction},
we show some generalization results from our method trained on Human3.6, considering
predictions on challenging images from Penn Action dataset.}

\subsection{Ablation studies}

In this part, we present additional experiments to provide insights
about our method and our design choices.

\noindent
\textbf{Network architecture}.
We evaluated three different network architectures as presented in
Table~\ref{tab:abla-netarch}. An off-the-shelf ResNet performed 62.2mm and
53.7mm, respectively when cut at blocks 4 and 5.  The proposed ResNet-U
improves on ResNet block 5 by 3.2mm while requiring 2.7M less parameters.

\begin{table}[]
  \centering
  \caption{
    \label{tab:abla-netarch}
    Evaluation of the network architecture, considering the backbone only
    (ResNet) cut at block 4 and block 5, and the refinement network (ResNet-U).
  }
\begin{tabular}{@{}cc|cc|cc@{}}
  \hline
    \multicolumn{2}{@{}c|}{ResNet block 4}%
    & \multicolumn{2}{c|}{ResNet block 5}%
    & \multicolumn{2}{c@{}}{ResNet-U} \\
                  MPJPE  & \small \#Par.%
                  & MPJPE  & \small \#Par.%
                  & MPJPE  & \small \#Par. \\ \hline
  62.2   & 10.5M  & 53.7 & 26M & 50.5     & 23.3M \\
  \hline
\end{tabular}
\end{table}

\noindent
\textbf{Absolute depth estimation}.
In Table~\ref{tab:abla-abs-depth}, we evaluate the influence of visual features
and bounding box position for the absolute depth estimation, considering the
mean root position error in mm (MRPE).  As can be observed, using only bounding
box features is insufficient to precisely predict the absolute $\eza{}$, but
when combined with visual features it further improves by 20mm,
which evidences the need of a global bounding box
information for that task.

\begin{table}[]
  \centering
  \caption{
    Absolute root joint position error in mm based on different features combinations.
    \label{tab:abla-abs-depth}
  }
  \begin{tabular}{@{}l|ccc@{}}
    \hline
    Features & Bounding box & CNN features & Combined \\
    \hline
    MRPE       & 375.4 & 100.1 & 80.1 \\ \hline
  \end{tabular}
\end{table}

\noindent
\textbf{The effect of multiple camera views}.
Since the proposed method predicts 3D poses in absolute camera coordinates and
is also capable of estimating the extrinsic camera parameters, we can use
multiple cameras to predict the same pose at inference time.
When considering multi-view scenarios, we can either use camera calibration,
when provided, or we can use our consensus-based optimization algorithm.

In Table~\ref{tab:abla-multiview} we present our results considering both 3D
pose estimation and absolute root position errors, as well as estimated and
ground truth camera parameters. We use multiple combinations of cameras, in
order to show the influence of different number of views.
As we can see, each camera lowers the error by about 5mm,
which is significant on Human3.6M.
We can also notice that our consensus optimization approach is capable
of providing highly precise estimation, even under uncalibrated conditions.

\begin{table}[hbtp]
  \centering
  \caption{
    \label{tab:abla-multiview}
    Results of our method on 3D human pose estimation and on root joint
    absolute error (MPJPE / MRPE) considering single and multi-view with
    different camera combinations.
  }
  \small
  \begin{tabular}{@{}l|cc|cc@{}}
    \hline
    \multirow{2}{*}{Method} & \multicolumn{2}{c|}{\small GT camera} & \multicolumn{2}{c@{}}{\small Estimated camera} \\
                            & \footnotesize MPJPE & \footnotesize MRPE & \footnotesize MPJPE & \footnotesize MRPE \\\hline
    Monocular          & 50.5 & 80.1 & -- & -- \\
    Monocular\footnotesize{ + h.flip} & \textbf{49.2} & \textbf{79.9} & -- & -- \\ \hline
    Cameras 1,2        & 45.7 & 73.3 & 52.2 & 167.0 \\
    Cameras 1,4        & 46.2 & 74.9 & 59.0 & 171.0 \\
    Cameras 1,2,3      & 41.8 & 57.4 & 47.9 & 143.8 \\
    Cameras 1,2,3,4    & \textbf{36.9} & \textbf{51.0} & \textbf{44.7} & \textbf{130.7} \\
    \hline
  \end{tabular}
\end{table}

\begin{figure}[hbtp]
  \centering
  \includegraphics[width=0.48\textwidth]{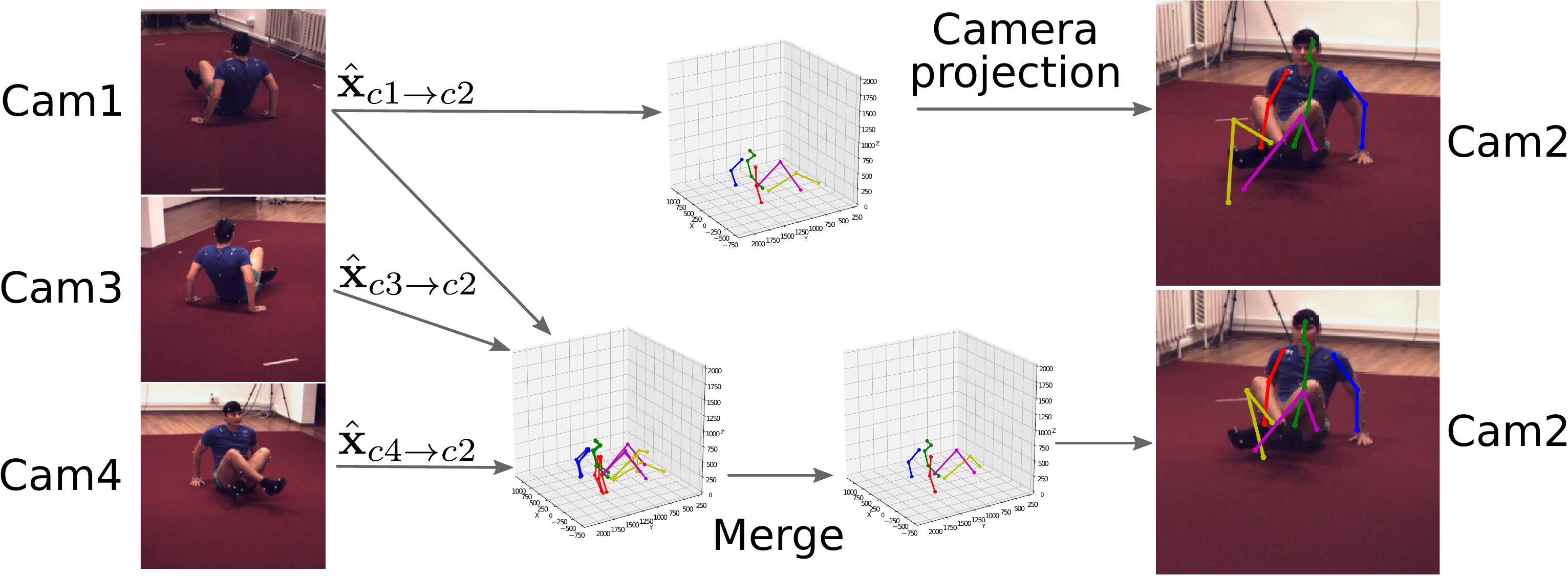}
  \caption{On top, the absolute prediction from camera 1 is projected into camera
  2 with considerable errors in occluded joints.  At the bottom,
  predictions from cameras 1, 3, 4 are projected into camera 2 and merged,
  improving the prediction significantly.}
  \label{fig:camera-projections}
\end{figure}

\begin{figure}[hbtp]
  \centering
  \includegraphics[width=0.45\textwidth]{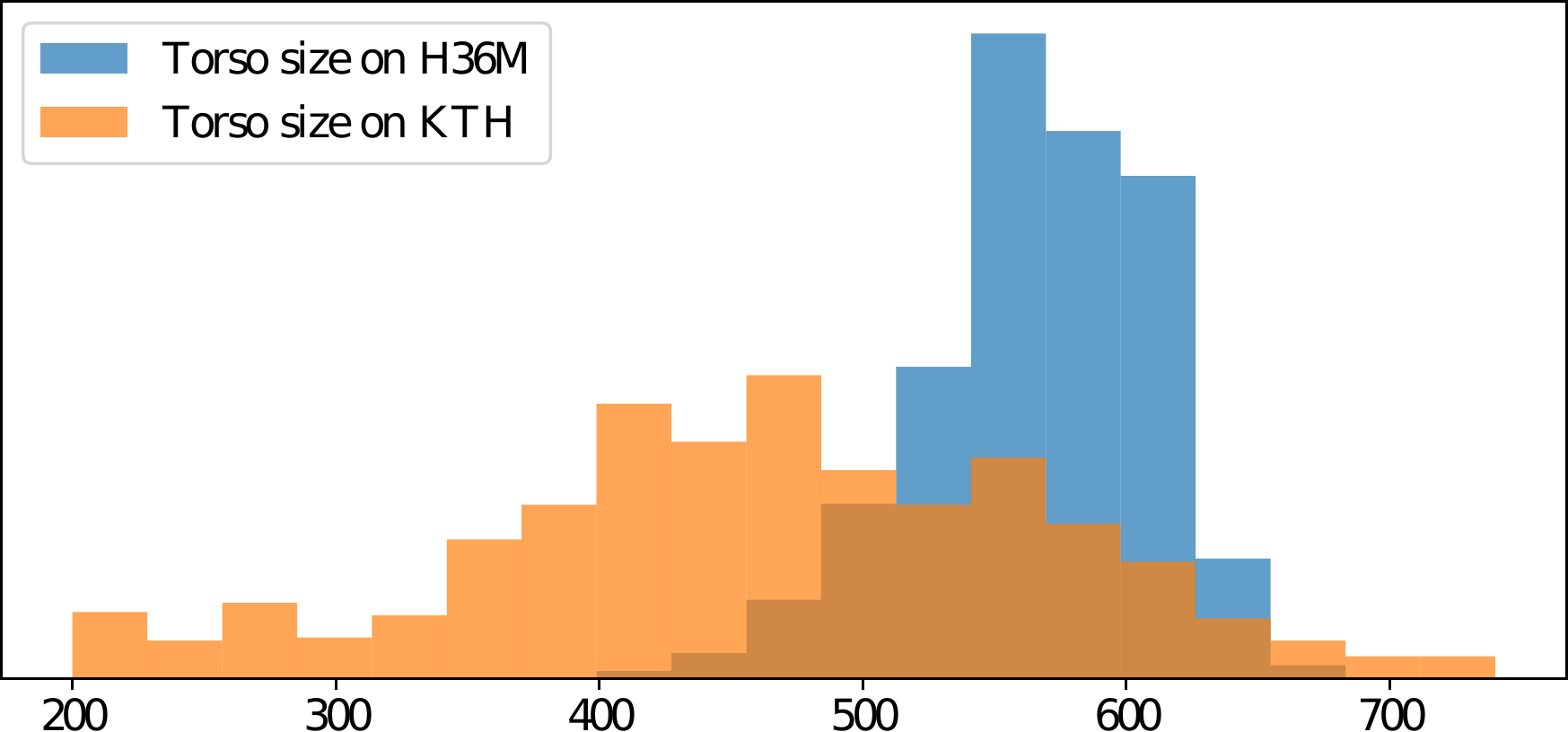}
  \caption{
  \revb{
  Torso size in mm of our estimated 3D poses on Human3.6M and on KTH, considering a zero-shot scenario.
  }
  }
  \label{fig:shift-kth}
\end{figure}

\begin{figure}[!h]
  \centering
  \includegraphics[width=0.14\textwidth]{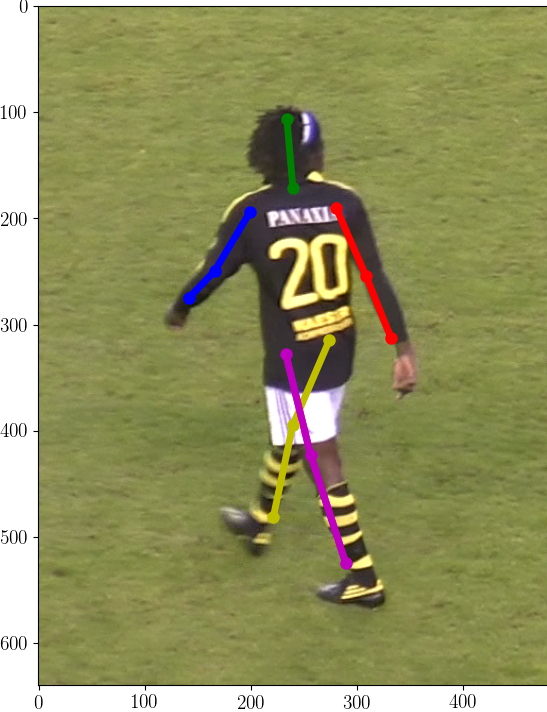}
  \includegraphics[width=0.14\textwidth]{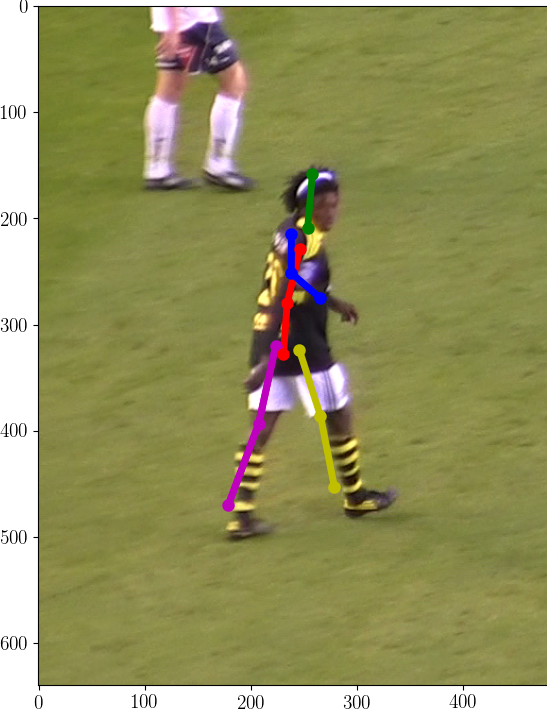}
  \includegraphics[width=0.14\textwidth]{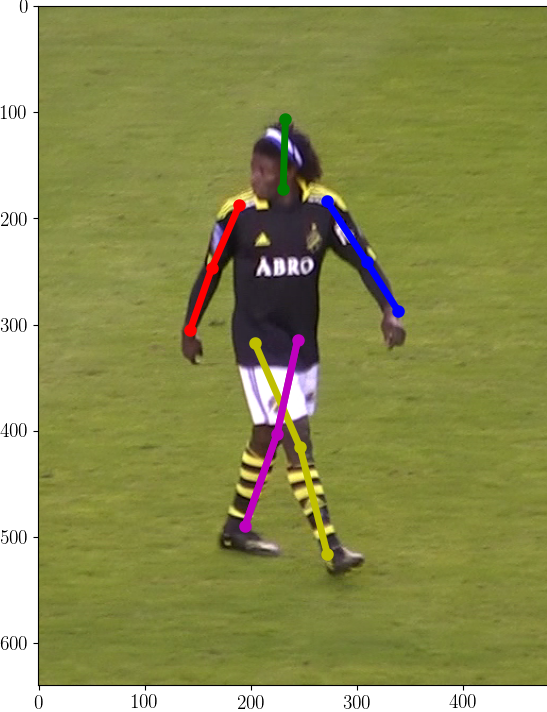}\\
  \includegraphics[width=0.14\textwidth]{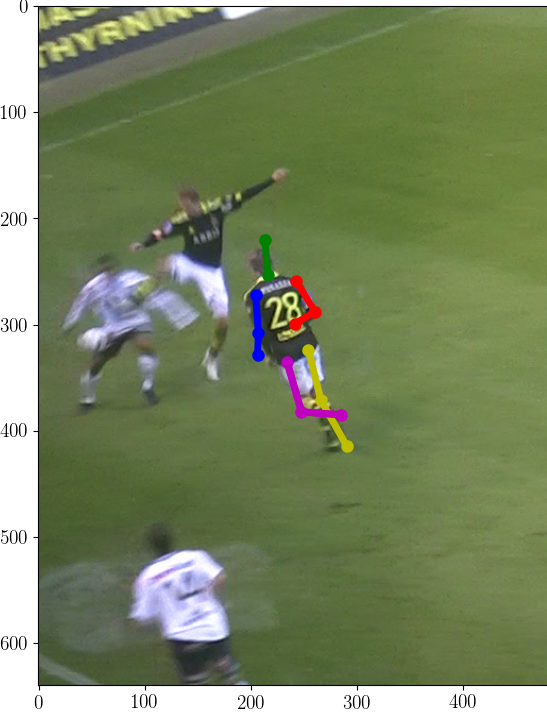}
  \includegraphics[width=0.14\textwidth]{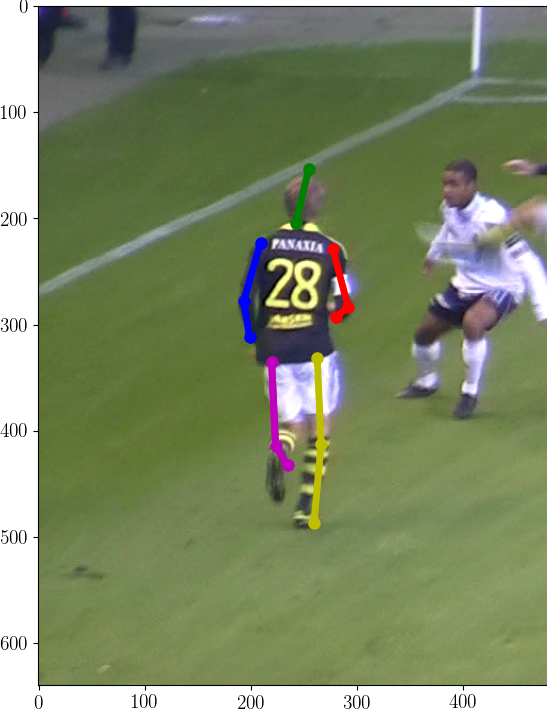}
  \includegraphics[width=0.14\textwidth]{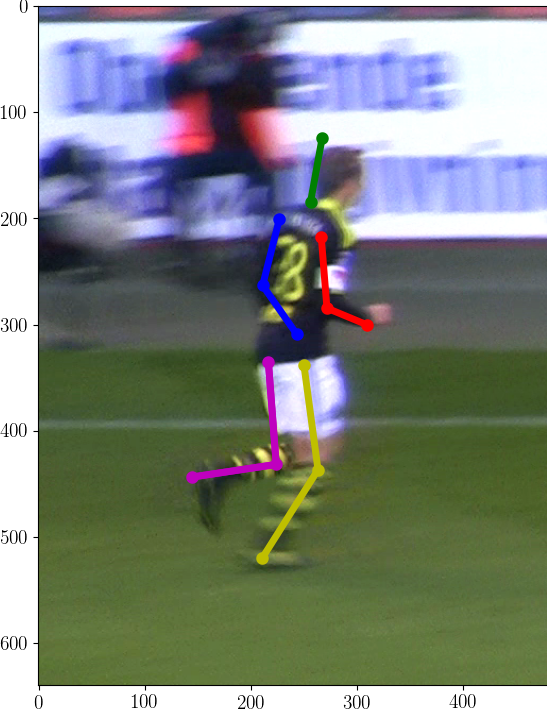}
  \caption{
  \revb{
    Qualitative results on KTH: projections of estimated 3D poses by our model trained on H36M.
    }
  }
  \label{fig:predictions_kth}
\end{figure}

\noindent
\revb{
\textbf{Zero-shot on new camera setup}.
We evaluated our method in a new camera setup considering a zero-shot scenario, where we used our model trained on Human3.6M and and evaluated it on KTH.
Due to the high disparity in the camera intrinsics between both datasets, the absolute depth predictions stayed in the range observed in Human3.6M, with an average of $4.040$ meters. The consensus-based algorithm still converged in the multi-view scenario, even though the final 3D poses are shifted to a smaller size due to the absolute depth predicted by our method (see Fig.~\ref{fig:shift-kth}).
To correct the scale and shift, we rescaled the predicted 3D poses using the torso size from KTH (the length from the neck to the hip center) and shifted our predictions to the KTH poses in the  hip center. After this, we computed the PCP metrics of our predictions, which results in $0.812$ and $.929$ for lower and upper legs, and in $0.620$ and $0.804$ for lower and upper arms.
Additional qualitative results on KTH are shown in Fig.~\ref{fig:predictions_kth}, where the final 3D estimations are also projected to the source images.
}

\revb{Finally}, Fig.~\ref{fig:camera-projections} shows an example of highly
occluded body parts where multiple camera predictions results in a
significantly better reconstruction. Note that in this case we are projecting
the estimated absolute 3D pose to a new point of view, not used during
inference. Despite the highly occluded joints in some views, the resulting
absolute pose is very coherent and has a reduced shift when our consensus-based
algorithm is used.

\subsection{\revb{Discussion}}

\revb{In the proposed approach, we have the advantage of predicting 3D human poses in absolute coordinates, which enables performing multi-view estimations while training the neural network model with monocular images. This aspect allows our method to be easily adapted to a variable number of cameras, without any additional training cost, as demonstrated in Table~\ref{tab:abla-multiview}. On the other hand, our method requires absolute 3D poses during training, which is a limiting factor, specially for training on datasets that provides only normalised human poses. The trained model can also be limited by the low variability of camera intrinsics during training, which may result in shift and scale deviations during inference on different cameras. As a future work, the proposed consensus-based optimization could be further integrated in the training pipeline, in order to allow a training process based on multiple views of the scene.}

\section{Conclusions}
\label{sec:conclusions}

In this paper, we have proposed a new method for the problem of predicting 3D
human poses in absolute coordinates and a new algorithm for multi-view
predictions optimization.
We show that, by casting the problem into a new perspective, we can benefit
from training with 2D and 3D data indistinguishably, while performing 3D
predictions in a more effective way. These improvements boost monocular 3D pose
estimation significantly. As another consequence of the absolute prediction, we
show that multi-view estimations can be easily performed from multiple absolute
monocular estimations, resulting in much higher precision than previous methods
in the literature, even when considering multiple uncalibrated images.




{
  \small
  \bibliographystyle{ieee}
  \bibliography{references}
}

\end{document}